\RequirePackage{amsmath}
\documentclass[runningheads]{llncs}
\usepackage{algorithmic}
\usepackage{algorithm}
\usepackage[T1]{fontenc}
\usepackage{graphicx}
 \usepackage[misc]{ifsym}
\usepackage{booktabs}
\usepackage[accsupp]{axessibility}
\usepackage{orcidlink}
\usepackage{cite}

\begin{document}
\title{Federated Learning with Flexible Architectures} 
%N.B.: Author information (both in the \author{} and \authorrunning{} command) should only be present in the Camera-Ready Version of your paper. The version that you initially submit for review, ought to be double-blind. So, when initially submitting your paper, use:
%\author{Author information scrubbed for double-blind reviewing}
\author{Jong-Ik Park \and Carlee Joe-Wong \Letter \orcidID{0000-0003-0785-9291}}
\authorrunning{Park and Joe-Wong}
\institute{Carnegie Mellon University, Pittsburgh PA 15213, USA \\
\email{cjoewong@andrew.cmu.edu}\\
\url{https://www.cmu.edu/}}

\maketitle              % typeset the header of the contribution
%
%%%%%
% Length 16 pages includes references and don't use negative vspace.
%%%%
\begin{abstract}
%The abstract should briefly summarize the contents of the paper in 150--250 words.
Traditional federated learning (FL) methods have limited support for clients with varying computational and communication abilities, leading to inefficiencies and potential inaccuracies in model training. This limitation hinders the widespread adoption of FL in diverse and resource-constrained environments, such as those with client devices ranging from powerful servers to mobile devices. 
To address this need, this paper introduces Federated Learning with Flexible Architectures (FedFA), an FL training algorithm that allows clients to train models of different widths and depths. Each client can select a network architecture suitable for its resources, with shallower and thinner networks requiring fewer computing resources for training.
Unlike prior work in this area, FedFA incorporates the \textit{layer grafting} technique to align clients' local architectures with the largest network architecture in the FL system during model aggregation.
Layer grafting ensures that all client contributions are uniformly integrated into the global model, thereby minimizing the risk of any individual client's data skewing the model's parameters disproportionately and introducing security benefits.
Moreover, FedFA introduces the \textit{scalable aggregation} method to manage scale variations in weights among different network architectures.
Experimentally, FedFA outperforms previous width and depth flexible aggregation strategies. Specifically, FedFA's testing accuracy matches (1.00 times) or is up to 1.16 times higher globally for IID settings, 0.98 to 1.13 times locally, and 0.95 times to 1.20 times higher globally for non-IID settings compared to earlier strategies. Furthermore, FedFA demonstrates increased robustness against performance degradation in backdoor attack scenarios compared to earlier strategies. Earlier strategies exhibit more drops in testing accuracy under attacks—for IID data by 1.01 to 2.11 times globally, and for non-IID data by 0.89 to 3.31 times locally, and 1.11 to 1.74 times globally, compared to FedFA. 
\keywords{Federated Learning  \and Heterogeneous Local Network Architectures \and Backdoor Attack}
\end{abstract}

\section{Introduction}
As the need for advanced decision-making capabilities in scenarios such as medical imaging and military operations grows, modern machine learning and deep learning techniques are increasingly in demand~\cite{antunes2022federated,demertzis2023federated}. To tackle potential privacy concerns associated with handling clients' data in these sensitive areas, federated learning (FL) has been developed. FL allows for training a shared machine learning model using data from multiple clients without the need to exchange or reveal their local data directly~\cite{mcmahan2017communication,lyu2020threats,nguyen2021federated}. Clients iteratively compute updates on local models and periodically synchronize these updates with an aggregation server to create a global model, which is sent back to the clients for another round of local model updates. 
However, \textit{traditional FL strategies struggle to integrate heterogeneous clients with varying computational and communication resources}~\cite{li2020review,diao2020heterofl,deng2022tailorfl}. 
For example, devices in a federated network may range from powerful fighter jets to less capable tanks in a military context or large hospitals to smaller clinics in healthcare. Prior works along these lines generally focus on reducing the number of local model updates that weaker clients complete during the training~\cite{ribero2020communication,ruan2021towards,park2022amble,li2020federated_prox}, which can slow model convergence. Moreover, each client must store a copy of and compute gradients on the full, possibly very large, model.

In this work, we explicitly account for heterogeneity in clients' compute and communication resources by allowing clients to customize their local model architectures according to their specific resources, ensuring efficient participation by avoiding delays from slower (straggler) clients, which could hinder the FL process and negatively impact global model updates~\cite{li2020federated,ruan2021towards}. Thus, our work falls into the same category as width-flexible FL aggregation strategies like HeteroFL~\cite{diao2020heterofl}, depth-flexible FL aggregation strategies such as FlexiFed~\cite{wang2023flexifed}, and strategies flexible in both width and depth like NeFL~\cite{kang2023nefl}. These strategies enable FL clients to train network architectures with variable depths and widths. 

\begin{figure}[t]
    \centering
    \includegraphics[width=\linewidth]{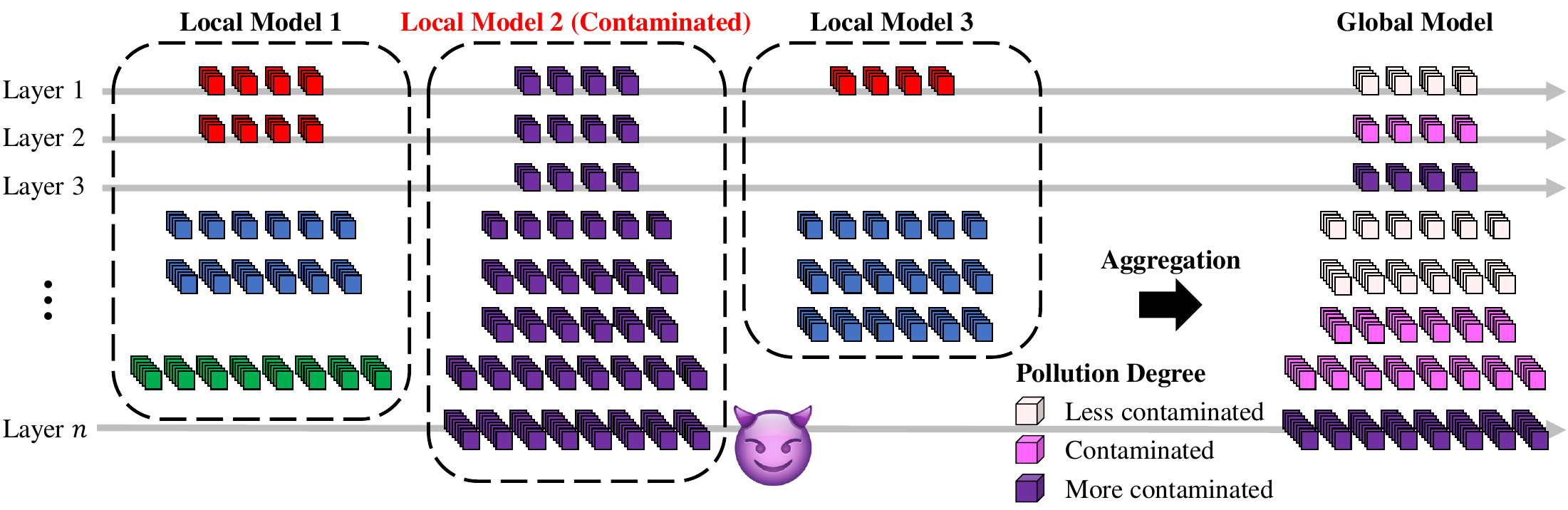}
    \caption{Aggregating heterogeneous networks introduces vulnerabilities due to incomplete aggregation and increased susceptibility to backdoor attacks for the global model.}
    \label{fig: model attack}
\end{figure}

However, prior works do not consider the fact that client networks' diversity (or heterogeneity) in FL presents \textbf{unique security challenges}. Combining models with different network architectures introduces weak points in the aggregation process that are susceptible to attacks~\cite{bagdasaryan2020backdoor,lyu2020threats}. These \textit{weak points} refer to weights that are \textit{incompletely aggregated}, since only a subset of clients compute their values due to the differences in network structures.
Attackers can exploit these vulnerabilities in commonly used \textit{backdoor attacks}~\cite{bagdasaryan2020backdoor,lyu2020threats}, which aim to induce inaccurate predictions on specific data inputs by manipulating model updates from malicious clients. By manipulating weights that are only updated by a few clients, attackers can successfully compromise the model, as depicted in the last two layers of the global model in Figure~\ref{fig: model attack}.

Following these concerns, another critical challenge arises from \textit{scale variations} in client weights due to the heterogeneous nature of network architectures~\cite{diao2020heterofl,vahidian2021personalized,deng2022tailorfl}. When clients possess varying numbers of layers and filters, scale variations arise, potentially causing unfairness in the global model aggregation: data from specific clients whose model weights have a larger scale may be disproportionately emphasized in the global model.

In response to these challenges, this paper proposes a novel strategy, \\ `\textbf{Federated Learning with Flexible Architectures}' (FedFA), that retains the benefits of employing heterogeneous model architectures while minimizing the impact of weak point attacks and addressing scale variation in aggregation. Our strategy aggregates model layers uniformly, regardless of the complexities of individual networks. 
We thus establish a global model that matches the greatest depth and width found among all local models.
This setup allows each client to contribute to the value of each weight in the global model, minimizing the risks associated with specific weak point attacks. Lastly, we propose a fair-scalable aggregation method to ensure fairness across local models and reduce the model bias from scale variations. 

In essence, our FedFA framework delivers four significant \textbf{contributions}: \\
\textbf{1)} We introduce a novel \textit{aggregation strategy} that is the first, to the best of our knowledge, to address security challenges in FL on heterogeneous architectures. FedFA uniformly incorporates layers from various client models into a unified global model, exploiting similarities between layers of a neural network induced by the common presence of skip connections.
\\
\textbf{2)} We are the first to effectively address scale variations in a dynamic training environment. We propose the \textit{scalable aggregation method} to compensate for scale variations in the weights of heterogeneous network architectures.
\\
\textbf{3) } FedFA \textit{utilizes NAS (Neural Architecture Search)}~\cite{li2023zico} to optimize each client's model architecture based on its specific data characteristics, thus elucidating the impact of employing optimal model architectures tailored to local data characteristics on both local and global model performance.
\\
\textbf{4) } In our experiments on Pre-ResNet, MobileNetV2, and EfficientNetV2, \textit{FedFA outperforms previous width- and depth-flexible strategies}. FedFA achieves accuracy improvements by factors of up to 1.16 in IID (independent and identically distributed) data settings and 1.20 in non-IID settings on the global model. In non-IID environments, clients' local accuracies increase by up to 1.13. Furthermore, FedFA demonstrates increased robustness. In contrast, prior strategies experienced accuracy declines under backdoor attacks by up to 2.11 in IID and up to 3.31 globally and 1.74 locally in non-IID settings compared to FedFA. Additionally, our experiments with a Transformer-based language model showed a significant reduction in perplexity, improving by 1.07 to 4.50 times.

We outline the `Related Work' in Section~\ref{sec:related work} and motivate FedFA in Section~\ref{sec:background}, which introduces previous width- and depth-flexible strategies and the model properties that we exploit. 
Next, `Flexible Federated Learning' in Section~\ref{sec:methodology} details the design of FedFA, and Section~\ref{sec:experimental evaluation} discusses our experimental results. We discuss directions of future work in Section~\ref{sec:future} and conclude in Section~\ref{sec:conclusion}.

\section{Related Work} \label{sec:related work}
\subsection{Heterogeneous Network Aggregation in Federated Learning}
\textbf{FlexiFed}~\cite{wang2023flexifed} is a depth-flexible strategy that aggregates common layers of clients' networks with varying depths, like in VGG-16 and VGG-19, forming global models from different layer clusters.
\textbf{HeteroFL}~\cite{diao2020heterofl} is a width-flexible strategy that accommodates clients with varying resources by aggregating networks of different widths. It selectively aggregates weights where available and employs a heuristic to manage weight variability~\cite{noci2023shaped}.
\textbf{NeFL}~\cite{kang2023nefl} combines width and depth flexibility, using skip connections to omit certain blocks and structured pruning for width control, similar to HeteroFL~\cite{diao2020heterofl}.
These strategies result in \textit{incomplete aggregation}, which poses security risks to the global model (see Figure~\ref{fig: model attack}). Here, \textit{incomplete aggregation} refers to the process where weights in layers or filters in the global model at a specific position are updated with contributions from only a subset of the participating local models rather than all.

Unlike HeteroFL, FlexiFed and NeFL do not consider scale variations between clients' model weights, leading to potential unfairness in model aggregation. Moreover, HeteroFL's scaling factors might be less relevant in architectures with batch normalization layers~\cite{ioffe2015batch}, which stabilize learning and reduce the need for additional scaling. For more on HeteroFL's scaling, refer to Appendix~\ref{apdx: HeteroFL}.
Furthermore, several other width- and depth-flexible strategies like Sub-FedAvg~\cite{vahidian2021personalized} and TailorFL~\cite{deng2022tailorfl} predominantly rely on online filter pruning, which can lead to significant computational overhead, contradicting their aim for computational efficiency. Therefore, in this study, we benchmarked our proposed FedFA strategy against HeteroFL, FlexiFed, and NeFL (see Section~\ref{sec:experimental evaluation}).

\subsection{Split Learning}
The \textbf{split learning} FL framework also allows clients to maintain a variable number of neural network layers, which connect to common layers stored at the aggregation server~\cite{pfeiffer2023federated,duan2022combined}. Clients can choose the number of local layers according to their computing resources and data characteristics~\cite{samikwa2022ares}. However, split learning requires intensive client-server communication, as clients cannot compute local model updates without communicating with the layers at the server~\cite{turina2021federated,han2021accelerating,oh2022locfedmix}.

\subsection{Skip Connections}
Modern network architectures have highlighted the significance of \textbf{skip connections} (or residual connections) in neural networks, a feature we utilize in our layer grafting method to mitigate the security risks of incomplete aggregation (e.g., ResNets~\cite{he2016deep}, MobileNets~\cite{howard2019searching}, and EfficientNets~\cite{tan2021efficientnetv2}). Skip connections allow gradients to bypass specific layers, mitigating vanishing gradients in deep learning~\cite{liu2020rethinking}. This functionality preserves training stability and enhances pattern recognition efficiency, making these networks suitable for resource-constrained devices~\cite{he2016deep,murshed2021machine}, and making layers similar~\cite{greff2016highway,veit2016residual,kang2023nefl} (See Appendix~\ref{apdx: layer grafting}.).
\section{Motivation: Challenges in Heterogeneous Aggregation} \label{sec:background}
\subsection{Security Concerns of Heterogeneous Network Aggregation}
Aggregating models from diverse network architectures presents security challenges, as malicious actors can exploit these strategies to carry out sophisticated and covert attacks, implanting subtle yet harmful alterations within model updates~\cite{bagdasaryan2020backdoor,tolpegin2020data}. These modifications, often in the form of triggers or slight changes, are designed to exploit the aggregation process covertly~\cite{abad2023sniper} and steer a model to degrade its accuracy or to embed hidden vulnerabilities, which become more pronounced over time~\cite{lyu2020threats,rodriguez2023survey}.
A particular point of vulnerability with heterogeneous client architectures is the layers that are not fully aggregated (i.e., \textit{incomplete aggregation}) in the global model (which has the largest width and depth across all local models) due to limited contributions from few clients, as depicted in Figure~\ref{fig: model attack}.

A common attack embeds a backdoor into a malicious client's local model in a heterogeneous FL setting. This hidden function or behavior, designed to remain dormant, activates only under specific conditions. Once integrated into the global model, these backdoors can trigger significant security breaches, such as targeted misclassifications~\cite{bagdasaryan2020backdoor,abad2023sniper}. Mathematically, a backdoor attack computes a malicious model update as follows:
\begin{equation}
\begin{aligned}
\Delta {M}_\text{malicious}^t \leftarrow \Delta {M}_c^t + \lambda \cdot \Delta {M}_{\text{backdoor}}
\end{aligned}
\label{eq: backdoor attack}
\end{equation}
Here, $\Delta M_c^t$ is the original update from client $c$ at global iteration $t$, and $\Delta M_{\text{backdoor}}$ represents the backdoor modification. $\lambda$ determines the intensity of the backdoor effect. The contribution of $\lambda \cdot \Delta M_{\text{backdoor}}$ to the aggregation process of all clients' model updates dictates the extent of damage to the global model. Specifically, weights of the global model that undergo incomplete aggregation are more susceptible to being compromised, as can be seen from the aggregation: 
\begin{align*}
\Delta M_G^t = \frac{1}{N} \left( \sum_{c=1}^{N-1} \Delta M_c^t + \Delta M_{\text{malicious}}^t \right) \not\approx \frac{1}{N-1} \sum_{c=1}^{N-1} \Delta M_c^t 
\end{align*}
For the global model update, \(\Delta M_G^t\), in the presence of many clients \(N\), the influence of an attack by malicious clients could be diluted. However, this dilution is limited to the weights that are updated by most or all clients.
Furthermore, by selecting the largest network architecture, attackers can amplify the effect of their attacks, in contrast to local clients who select network architectures based on their resource capabilities or the characteristics of their local data.

\subsection{Scale Variations in Heterogeneous Networks}
During the training process, the gradients of weights can vary significantly across different network architectures, influenced by the number of weights present in each network before applying the loss function. As a result, the magnitudes of weight changes during each step of gradient descent (i.e., step sizes) can differ due to the variations in gradients during optimization. This leads to scale variations across the heterogeneous networks~\cite{hanin2018neural} (refer to Appendix~\ref{apdx: scale variation} for more details).
In FL environments, such variations significantly impact the performance and accuracy of the aggregated global model. For instance, consider two client models within an FL system, Model A and Model B; each has a distinct architecture. The magnitude of their updates in a given round of FL, $\Delta M_A$ and $\Delta M_B$, can differ significantly based on their respective models' complexities.
When these models are aggregated using an unweighted averaging method, as is typical in federated learning, we have: 
\[\Delta{M} = \frac{1}{2} (\Delta M_{A} + \Delta M_{B})\]
This may lead to an imbalance, causing the aggregated model update, $\Delta{M}$, to be skewed towards the model with a larger magnitude of weights.
\section{Flexible Federated Learning} \label{sec:methodology}
\begin{figure*}[!ht]
    \centering
    \includegraphics[width=\linewidth]{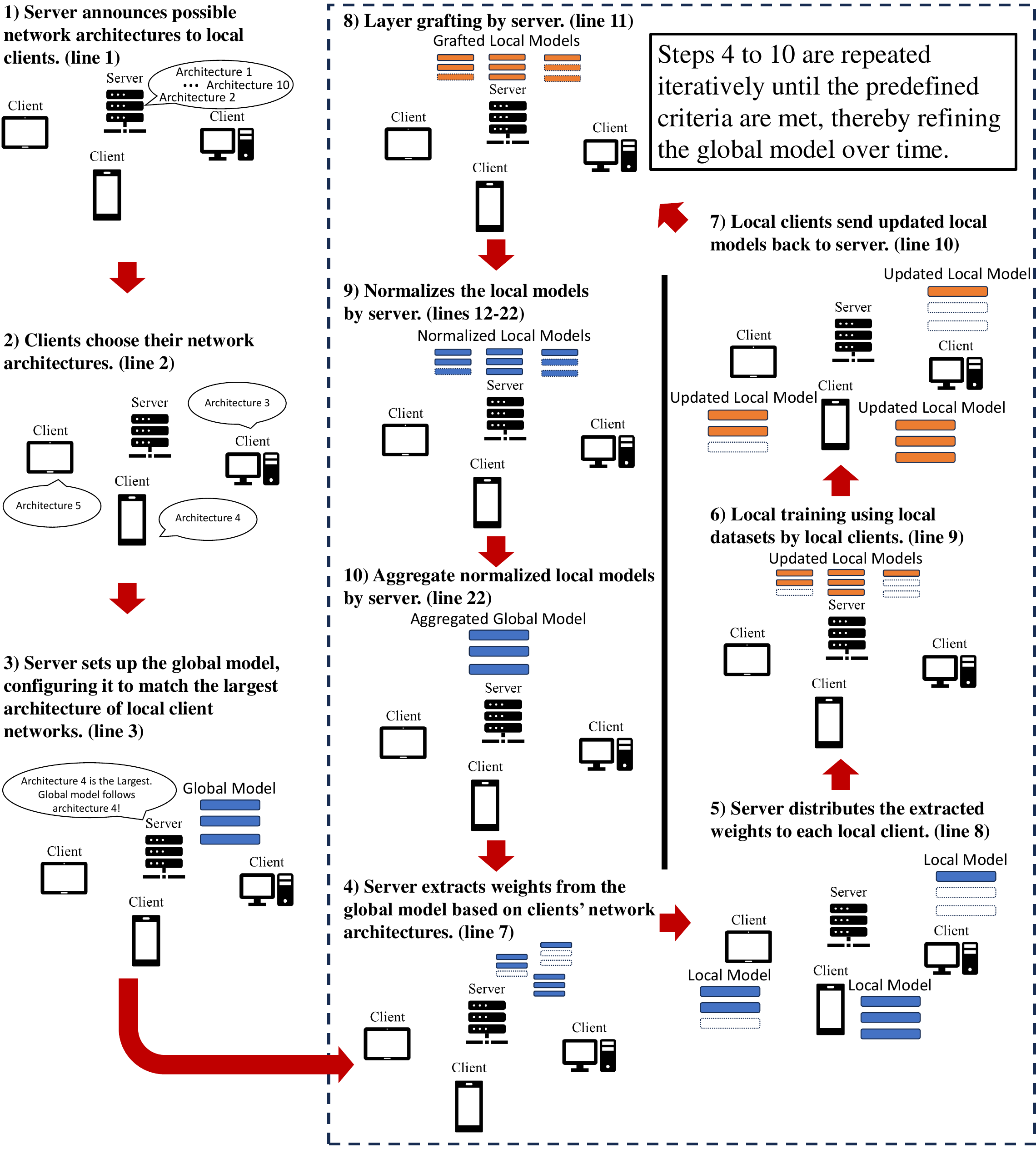}
    \caption{The FedFA workflow: Server announces network architectures, clients select and send preferences, server configures the global model, clients perform local training, updates are sent to the server, where they are grafted, normalized, and aggregated, iterating until convergence criteria are achieved. Each step in the workflow is mapped to specific lines in Algorithm~\ref{algorithm: FedFA} and depicted in the corresponding steps of the figure.}
    \label{fig: diagram}
\end{figure*}
This section details our methodology for implementing Federated Learning with Flexible Architectures (FedFA). We first introduce an overview of our FedFA procedure and then focus on the procedure for the layer grafting method, which ensures security by enforcing complete aggregation. Additionally, we introduce scaling factors for normalizing model weights, a critical component for achieving fair aggregation within the FedFA framework.

\subsection{FedFA Procedure}
\begin{algorithm}[!ht]
\caption{FedFA with the layer grafting and the scalable aggregation methods. The algorithm operates over $T$ rounds with a client set $\mathcal{C}$. Here, the clients' participating rate is $C$. Each client $c \in \mathcal{C}$ selects a model architecture from a predefined set $\mathcal{A}$. The server determines the maximal architecture width $N_{width,\max}^{(l)}$ and depth $N_{depth,\max}^{(s)}$. The global model, $M_G^t$, is updated at the server through the aggregation of client updates.}
\begin{algorithmic}[1]
\label{algorithm: FedFA}
\REQUIRE Local datasets $\mathcal{D} = \{D_c|c \in \mathcal{C}\}$.
\ENSURE Updated global model $M_G^T$.
\STATE Server proposes architecture set $\mathcal{A}$.
\STATE Clients select network architectures from $\mathcal{A}$ using NAS methods and report their architectures (width $N_{width,c}^{(l)}$ and depth $N_{depth,c}^{(s)}$) to the server.
\STATE Initialize the global model, $M_G^0$, with $N_{width,\max}^{(l)}$, $N_{depth,\max}^{(s)}$. \COMMENT{Server}
\FOR{$t=0$ to $T$}
    \STATE Select a subset $\mathcal{C}_{sel}$ of $m = C \times |\mathcal{C}|$ clients. \COMMENT{Server}
    \FOR{all clients $c$ in $\mathcal{C}_{sel}$} 
        \STATE Extract $M_c^t$ from $M_G^t$ according to $N_{width,c}^{(l)}$ and $N_{depth,c}^{(s)}$. \COMMENT{Server}
        \STATE Distribute $M_G^t$ to the client $c$. \COMMENT{Server}
        \STATE $M_c^{t+1} \gets \text{LocalUpdate}(M_c^t, D_c)$ \COMMENT{Client $c$}
        \STATE Send $M_c^{t+1}$ to the server. \COMMENT{Client $c$}
        \STATE Apply the layer grafting to $M_c^{t+1}$. \COMMENT{Server, Algorithm~\ref{algorithm: layer grafting}}
        \STATE $M_{95\%, c}$ $\gets$ Under 95th percentile values of $M_c^{t+1}$  for each layer \COMMENT{Server}
    \ENDFOR
    \FORALL{layers $l$ in $M_G^t$}
            \STATE $M_G^{'(l)}, \gamma^{(l)} \gets $ Zeros($M_G^{t, (l)}$) \COMMENT{Server}
            \FOR{all clients $c$ in $\mathcal{C}_{sel}$} 
                {\scriptsize
                \STATE$C_{I},C_{o}$} $\gets $ Input and output channel sizes of $M_c^{(l)}$ \COMMENT{Server}
                \STATE $\alpha_c^{(l)} \gets \frac{\frac{1}{m}\sum_{\kappa \in \mathcal{C}_{sel}} ||M_{95\%,\kappa}^{(l)}||}{||M_{95\%, c}^{(l)}||}$ \COMMENT{Server}
                \STATE Add  $N_{D_c} \alpha_c^{(l)} M_c^{(l)}$ to $M_G^{'(l)}[:C_{o},:C_{I}]$. \COMMENT{Server}
                \STATE Add $N_{D_c}$ to all elements of $\gamma^{(l)}[:C_{o},:C_{I}]$. \COMMENT{Server}
                % \carlee{$D_c$ is the local dataset. Why does the server have access to it?} 
            \ENDFOR
        \STATE $M_G^{(l)} \gets  M_G^{'(l)}/\gamma^{(l)}$ \COMMENT{Server}
    \ENDFOR
    \STATE $M_G^{t+1} \gets M_G^t$ \COMMENT{Server}
\ENDFOR
\newline
\STATE \textbf{Function} LocalUpdate{$(M, D)$}
\STATE $\;\;\;$ Update model $M$ using local dataset $D$.
\STATE \textbf{Return} Updated model $M$ 
\end{algorithmic}
\end{algorithm}

The FedFA algorithm, as presented in Algorithm~\ref{algorithm: FedFA}, incorporates the layer grafting and the scalable aggregation methods into the FL paradigm. Initially, the server proposes a variety of network architectures (line 1). Clients then choose their architectures, e.g., using Neural Architecture Search (NAS) methods (line 2), followed by the server setting up the global model with the maximum possible number of weights (line 3). The algorithm begins its iterative process by selecting a random subset of clients to update the model in each round (lines 5-9).
Subsequently, the layer grafting method (explained in Section~\ref{subsec : layer grafting}) will adjust each client's updated local model to align with the global model's architecture (line 11). 

For scalable aggregation, the server first finds the weights below the 95th percentile in each layer of each local model from each participating client (line 12).  
With these extracted weights, which exclude outliers, the server calculates the scaling factor, $\alpha_{c}^{(l)}$, for each layer (line 18).
Normalization of each local model and aggregation of these normalized local models then ensures that the updates from all participating clients are aggregated in a balanced manner, preserving uniformity and scale consistency throughout the network (lines 16-22) (more details are in Section~\ref{subsec : scalable aggregation}). 

Since local models might differ in width from the global model, contiguous structured pruning~\cite{diao2020heterofl,kang2023nefl} is used (line 19). This process accumulates the weights of local models only at the same position, considering their input and output channel sizes ($[:C_O, :C_I]$). Here, $M_G^{'(l)}$ and $\gamma^{(l)}$ are temporary placeholders with the same architecture as the global models, initialized with zeros for all elements in every round (line 14). $M_G^{'(l)}$ is used for accumulating local updates (line 19), and $\gamma^{(l)}$ is for the weighted average of these local updates (line 20). $\gamma^{(l)}$ considers the number of data samples for each client of line 1, $N_{D_c}$, aligning with the original FedAvg algorithm~\cite{mcmahan2017communication}. If the server cannot even access the number of data samples, we take $N_{D_c} = 1$. After accumulation, the server can obtain the updated global model (line 24) by element-wise dividing $M_G^{'(l)}$ by $\gamma^{(l)}$ for every layer $l$ (line 22). The algorithm continues through these rounds until predefined criteria are met. The overall FedFA process is visually summarized in Figure~\ref{fig: diagram}. We also show the effectiveness of heterogeneous network aggregation strategies in Appendix~\ref{apdx: heteroaggre effectiveness} and the convergence analysis of FedFA in Appendix~\ref{apdx: convergence analysis}.

\subsection{Layer Grafting Method for Ensuring Security}
\label{subsec : layer grafting}
In FL, client models can vary in architecture due to differences in computational resources and data characteristics. This heterogeneity can lead to adversarial attacks during the aggregation of local models, potentially compromising the security of the global model. To address these issues, we introduce the layer grafting method (line 11 in Algorithm~\ref{algorithm: FedFA}), which ensures uniformity in model architectures while accommodating client-specific characteristics.

\begin{algorithm}[ht]
\caption{The Layer Grafting method. This algorithm standardizes the depth of each section $M_c^{(s)}$ in local model $M_c$ to the maximum depth $N_{depth,\max}^{(s)}$ across all clients $c$. $N_{depth,c}^{(s)}$ denotes the current depth of section $M_c^{(s)}$, while $\Delta D$ represents the depth difference that needs to be augmented. The last residual block in a section is denoted by $R_{last}^{(s)}$, and $\oplus$ symbolizes the grafting operation of this block to the section. The process iteratively augments each section's depth $N_{depth,c}^{(s)}$ by grafting $R_{last}^{(s)}$ until the target depth $N_{depth,\max}^{(s)}$ is matched.}
\label{algorithm: layer grafting}
\begin{algorithmic}[1]
\REQUIRE Local model $M_c$ for client $c$, maximum depth $N_{depth,\max}^{(s)}$ for each section $M_c^{(s)}$ across all clients.
\ENSURE Updated model $M_c$ with each section $M_c^{(s)}$ augmented to depth $N_{depth,\max}^{(s)}$.
\FOR{each section $M_c^{(s)}$ in model $M_c$}
    \STATE $N_{depth,c}^{(s)} \gets \text{current depth of section } M_c^{(s)}$
    \STATE $\Delta D \gets N_{depth,\max}^{(s)} - N_{depth,c}^{(s)}$
    \IF{$\Delta D > 0$}
        \STATE $R_{last}^{(s)} \gets \text{last residual block in section } M_c^{(s)}$
        \FOR{$d = 1$ to $\Delta D$}
            \STATE $M_c^{(s)} \gets M_c^{(s)} \oplus R_{last}^{(s)}$
        \ENDFOR
    \ENDIF
\ENDFOR
\end{algorithmic}
\end{algorithm}

The layer grafting method, as described in Algorithm~\ref{algorithm: layer grafting} and illustrated in Figure~\ref{figure: layer grafting} in the Appendix, aims to standardize the depth of each section across the FL client models. In this context, a `section' is a part of the model where residual blocks share the same sequence of filter numbers in layers. A single model may comprise multiple such sections, each containing several residual blocks.

This addition is iteratively performed until the section reaches the specified maximum depth (lines 4-9 of the Algorithm~\ref{algorithm: layer grafting}). This systematic addition of residual blocks guarantees a consistent depth across all client models, thereby preserving architectural coherence within the FL network. Further details and the rationale behind layer grafting, particularly regarding the similarity of layers within residual blocks, are elaborated in Appendix~\ref{apdx: layer grafting}.

To see how layer grafting mitigates the potential risk of backdoor or poisoning attacks in aggregations across heterogeneous client architectures~\cite{diao2020heterofl,wang2023flexifed,kang2023nefl}, we examine the aggregation for the global FL model with layer grafting, assuming the commonly used averaging method~\cite{mcmahan2017communication}: 
\begin{align*}
M^{t}_G 
&=\frac{1}{N} \left(\sum_{c=1}^{N-1}(M^{t-1}_G+ \Delta {M}_c^t)+ (M^{t-1}_G+\Delta {M}_{\text{malicious}}^t)\right) \\
&\approx
\frac{1}{N-1} \sum_{c=1}^{N-1} \left( M^{t-1}_G+ \Delta {M}_c^t \right)
\end{align*}
In this simple aggregation formula~\cite{mcmahan2017communication}, $M^{t}_G$ and $M^{t-1}_G$ represent the global models at iterations $t$ and $t-1$, respectively. $N$ denotes the total number of clients, and $\Delta {M}_c^t$ are the updates from each individual client $c$. This equation illustrates how the influence of malicious updates is diluted in a complete aggregation, especially for a large number ($N$) of clients. This mitigation is effective if the updated weights,  $M^{t-1}_G+\Delta M_c^t$ and $M^{t-1}_G+\Delta M_{\text{malicious}}^t$ are on the same scale.

\subsection{Scalable Aggregation: Normalization of Local Model Weights}
\label{subsec : scalable aggregation}
In addressing scale variations stemming from heterogeneous network architectures in FL, we incorporate a crucial normalization step in the model aggregation process (lines 18-19 in Algorithm~\ref{algorithm: FedFA}). This involves applying scaling factors to the weight updates from each client model to ensure balanced contributions in the aggregated model.

The scaling factor, denoted as $\alpha_{c}^{(l)}$ for layer $l$ of each local model, is calculated in response to the diverse scales of weight updates from different network architectures. These factors are determined based on the L2 norm to prevent larger updates from disproportionately influencing the global model.

The formula for the scaled weights in the FedFA framework is:
\[
\alpha_c^{(l)} M_c^{(l)} = \frac{\frac{1}{m}\sum_{\kappa \in \mathcal{C}_{sel}} ||M_{95\%,\kappa}^{(l)}||}{||M_{95\%, c}^{(l)}||} M_c^{(l)} \text{ where } c \in \mathcal{C}_{sel}
\]
Here, $M_c^{(l)}$ represents the weights from the local model of client c for layer $l$. In one aggregation round, $m$ denotes the number of participating clients, $\mathcal{C}_{sel}$. The L2 norm $||M_{95\%, \kappa}^{(l)}||$ refers to the weights within the inner part of the 95th percentile from client $\kappa$ for the same layer $l$. 
Similarly, $||M_{95\%, c}^{(l)}||$ signifies the weights under the 95th percentile in layer $l$ for client $c$. We utilize the 95th percentile as it effectively mitigates the impact of outliers, which could otherwise skew the accuracy of scale calculations. This approach is beneficial in reducing the influence of anomalous weight values that may arise from noisy data or atypical client models.

This normalization process, with the scaling factor $\frac{1}{||M_{95\%, A}^{(l)}||}$, applied layer-wise, ensures that the aggregated model accurately reflects the diverse architectures in the network. 
Furthermore, the averaging component, represented by the numerator $\frac{1}{m}\sum_{\kappa \in \mathcal{C}{sel}} ||M_{95\%,\kappa}^{(l)}||$, moderates the convergence speed by averaging the magnitude of updates across the participating clients. This leads to a more balanced and representative global model, adjusting for scale variations across different client models and enhancing the overall fairness of the FL system.

\subsection{Global Model Distribution Step}
The model distribution step, detailed in Algorithm~\ref{algorithm: Model Distribution} in the Appendix, focuses on tailoring the aggregated global model to align with the unique architectural requirements of each client (line 7 in Algorithm~\ref{algorithm: FedFA}). This critical process involves modifying the global model to conform to each client's model's specific depth and width weights. To achieve this, the algorithm systematically adjusts the global model by reducing its depth (lines 3-6 in Algorithm~\ref{algorithm: Model Distribution}) and width (lines 8-11 in Algorithm~\ref{algorithm: Model Distribution}) to those specified by each client's architecture. %This reduction is meticulously executed to ensure that the dimensions of the global model precisely match the predetermined architectures unique to each client. 
By following this procedure, the global model is effectively customized, making it compatible with the diverse architectures of all participating clients in the FL network.
\begin{table*}[t]
    \centering
    \caption{Global and local testing accuracies for Pre-ResNet, MobileNetV2, and EfficientNetV2 on CIFAR-10, CIFAR-100, and Fashion MNIST datasets in IID and non-IID scenarios. FedFA's depth-, width-, and both depth and width-flexible approaches lead to higher test accuracy with no malicious clients and lower accuracy drops (under attacks with intensity $\lambda = 20$ and 20\% malicious clients) compared to prior depth-, width-, and both depth and width-flexible approaches. The maximum and minimum test accuracies are highlighted in bold for each scenario.}
    \includegraphics[width=\linewidth]{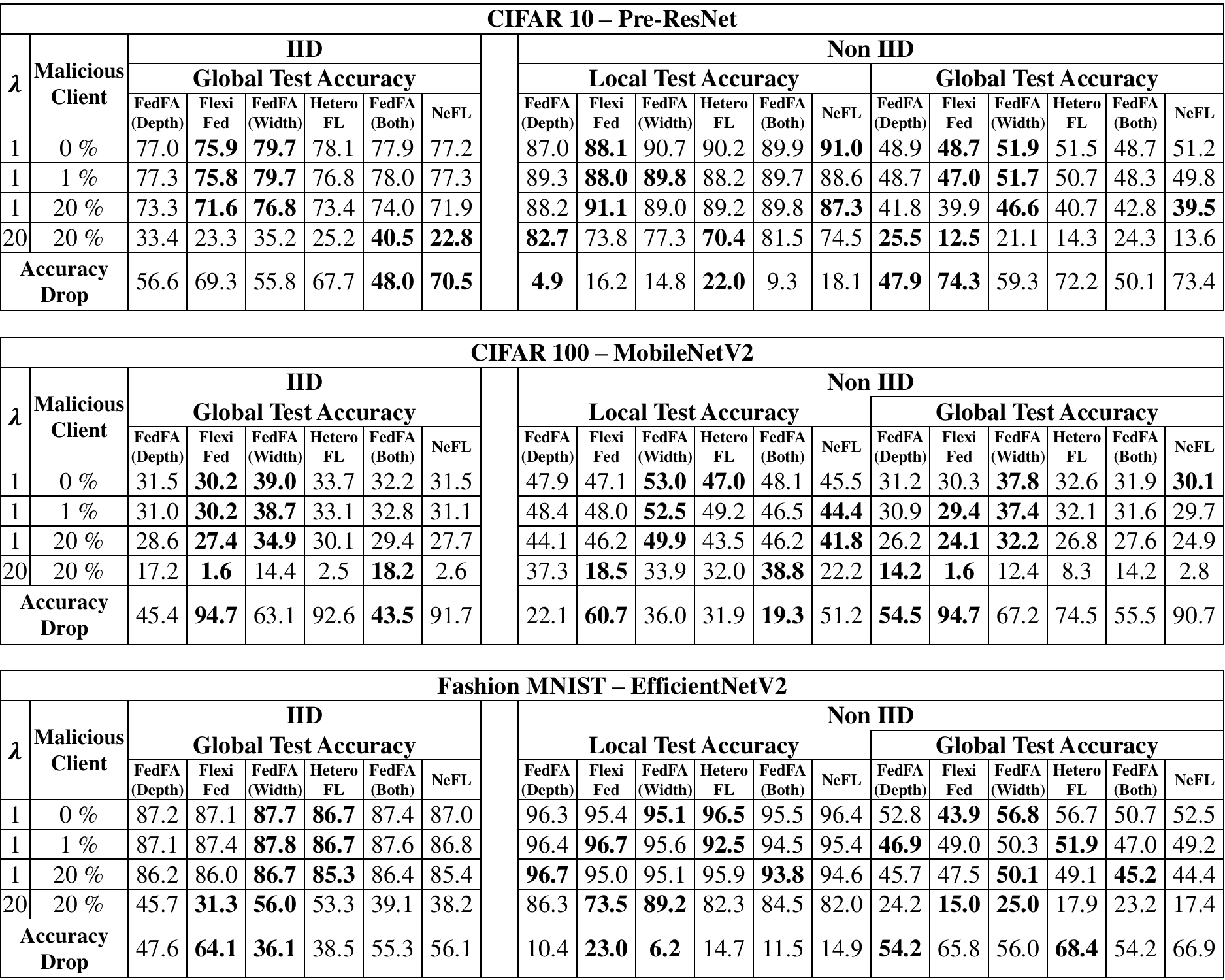}
    \label{table: test results}
\end{table*}
\section{Experimental Evaluation} \label{sec:experimental evaluation}
This section evaluates FedFA's performance by benchmarking its testing accuracy, robustness against backdoor attacks, and computational complexities. This evaluation uses local and global test datasets, comparing FedFA with previous aggregation strategies offering width and depth flexibility.
\subsection{Experimental Setup}
We assess Pre-ResNet, MobileNetV2, and EfficientNetV2 using the CIFAR-10, CIFAR-100, and Fashion MNIST datasets in IID and non-IID environments. In IID settings, each client has samples from all classes, with a uniform data distribution where the minimum number of samples for a client can be up to half the maximum number of samples for any other client.

For Non-IID settings, clients get samples from 20\% of the dataset classes but maintain equal samples for each class they hold. Here, during local training, clients zero-out logits for absent classes. We replace typical batch normalization layers with static versions, as seen in HeteroFL~\cite{diao2020heterofl}. We also utilized a language model with a Transformer using WikiText-2 to demonstrate that our method is generalizable. Detailed network structures are presented in Table~\ref{figure: model archi 1}, and more training details are in Table~\ref{figure: test condition} in the Appendix.

\begin{figure*}[t]
    \centering 
    \includegraphics[width=\linewidth]{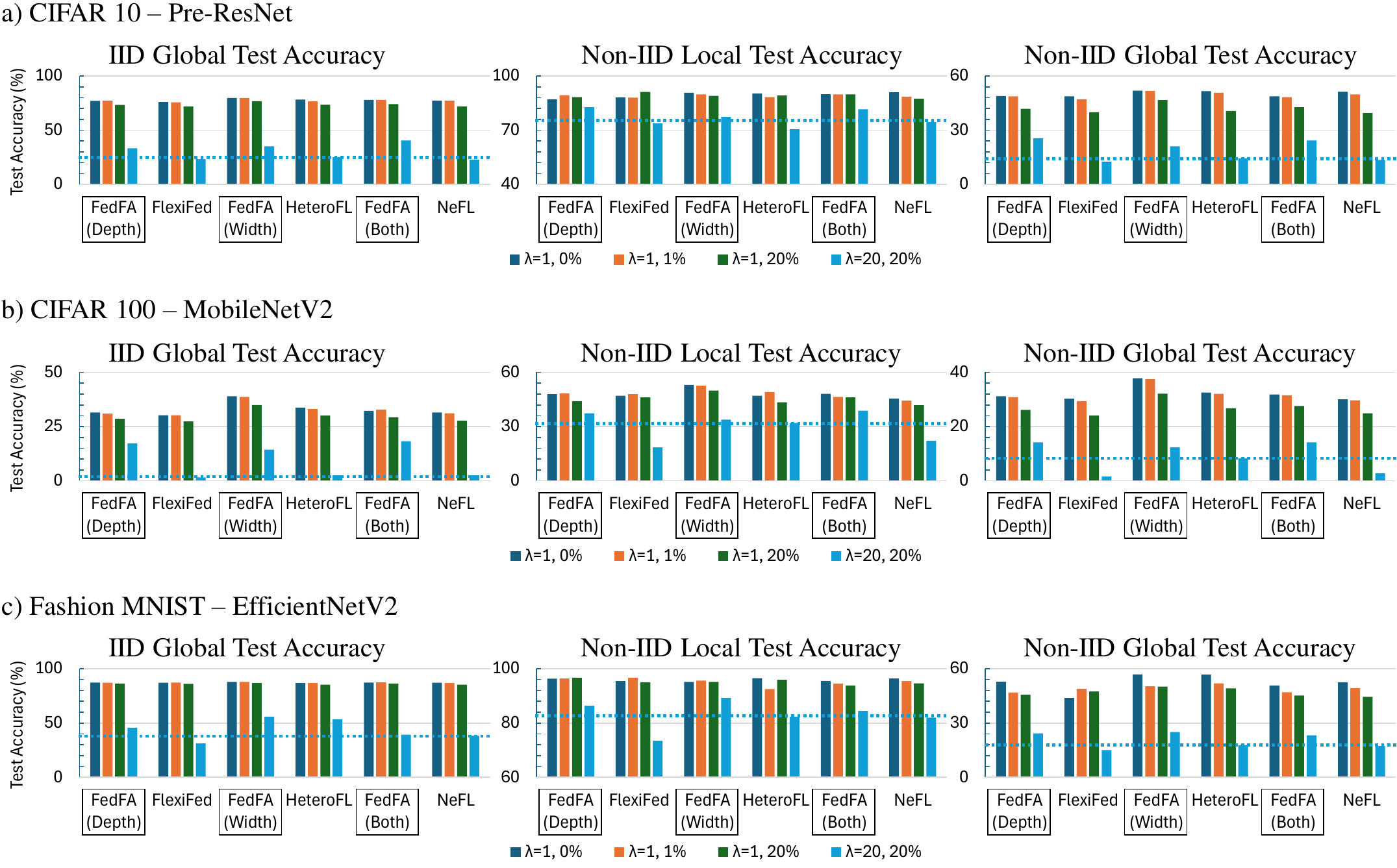}
        \caption{Visualizations of FedFA's robustness against backdoor attacks in different FL settings across the CIFAR-10 with Pre-ResNet, CIFAR-100 with MobileNetV2, and Fashion MNIST with EfficientNetV2 datasets. The blue dotted lines are positioned below the lowest accuracy of FedFA but above the next highest accuracy among FlexiFed, HeteroFL, and NeFL. They underscore the robustness of FedFA when the attack intensity \(\lambda=20\) with 20\% malicious clients.}
    \label{figure: graph teest results}
\end{figure*}

\begin{table*}[t]
    \centering
    \caption{Computational complexities for Pre-ResNet, MobileNetV2, and EfficientNetV2 on CIFAR-10, CIFAR-100, and Fashion MNIST datasets in IID and non-IID scenarios. FedFA shows comparable complexity to its baselines.}
    \includegraphics[width=\linewidth]{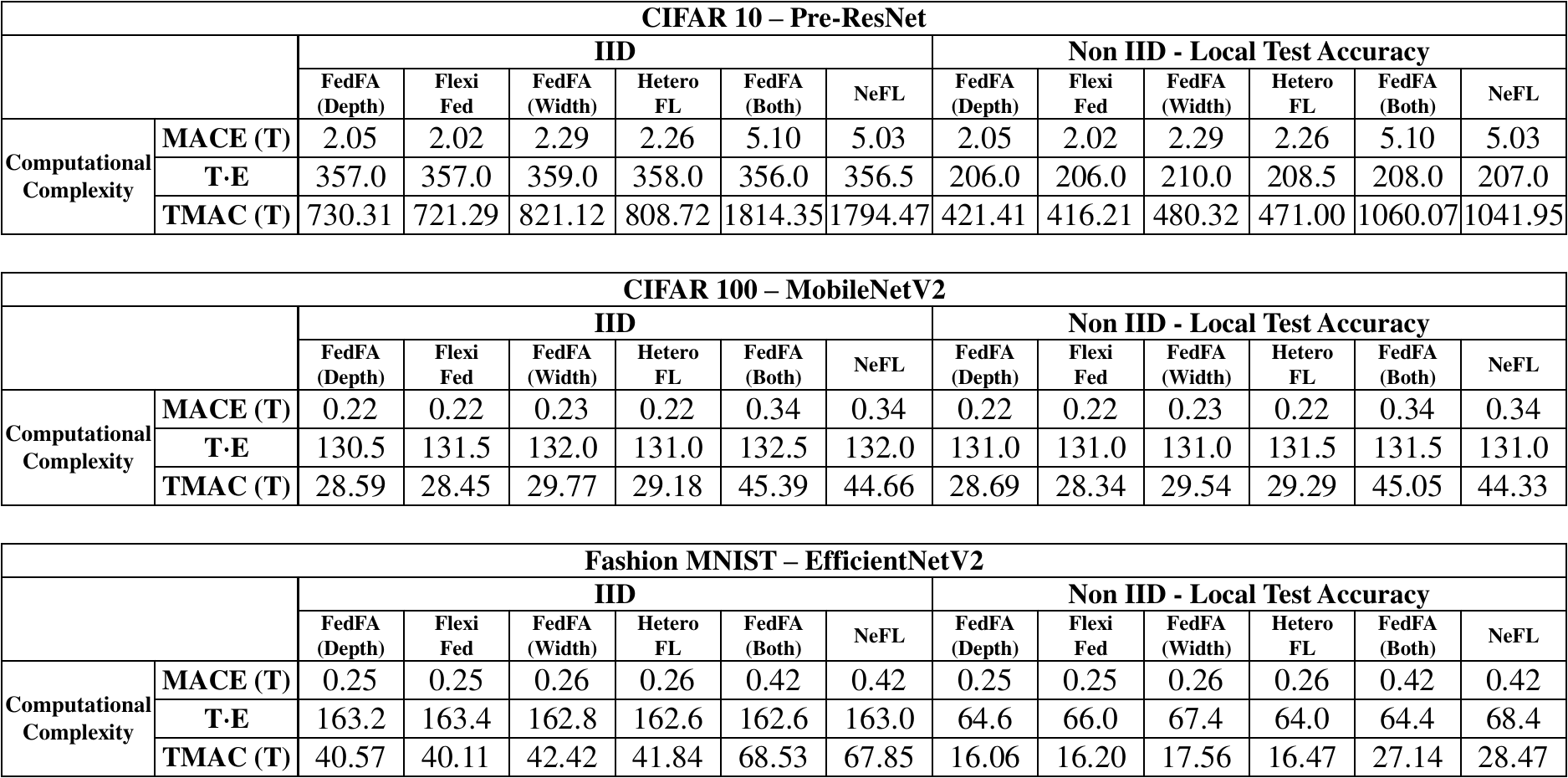}
    \label{table: computation results}
\end{table*}

Evaluations were conducted under four scenarios with varying impacts of backdoor attacks from malicious clients. Here, the backdoor attacks involve the random shuffling of the data labels among clients to induce misclassification. Scenarios have different portions of malicious clients over entire local clients (0 \%, 1 \%, and 20 \%) and two intensities of attacks, (\(\lambda=1, 20\) in Eq.~\ref{eq: backdoor attack}). Also, for all scenarios, we assume that half of the clients have limited computational resources and choose the smallest architectures. The other clients choose their architectures employing ZiCo~\cite{li2023zico}, a cost-effective NAS method that requires only forward passes and uses an evolutionary algorithm. This method decides local network architectures among the network candidates specified in Table~\ref{figure: model archi 2} in the Appendix, based on local data for each client.

\subsection{Baselines and Metrics}
After aggregating the local models, we calculate the \textit{global testing accuracy} using a global test dataset. In non-IID settings, we additionally use several local test datasets extracted from the global dataset, ensuring they reflect the local clients' class distributions. After local training and before aggregation, we test the local models to determine an \textit{average local testing accuracy}. This metric allows us to measure the effectiveness of local personalization. For the Transformer model, we use \textit{average local perplexity} to assess the performance of the local clients' language models after local training for every round. Here, perplexity measures how well a probability model predicts a sample. 

To evaluate \textit{computational complexity}, we rely on multiply-accumulate (MAC) calculations. $\text{MACS}_{n=i}$ is the MAC for one local epoch of a local model with given local data, differentiated by architecture $n=i$. $N_{n=i}$ counts such architectures in the FL system. The average MAC is $\overline{\text{MACS}}$ given varied complexities among local models. The MAC is $\text{MACE}=N_p \cdot \overline{\text{MACS}}$ for a single local epoch with all clients. The FL system's total MAC, $\text{TMAC}$, is found by multiplying total rounds of aggregation steps, $T$, by local epochs, $E$. 
\subsection{Evaluation}
\subsubsection{Testing Performance and Robustness Against Backdoor Attacks} 
Table~\ref{table: test results} shows the testing results of scenarios with varying model depths and widths for FedFA and previous flexible aggregation strategies in width and depth. Each scenario was tested three times, and the table presents the average results.

When only depth is varied, \textit{FedFA outperforms FlexiFed} with a 1.00 (equivalent) to 1.04 times improvement globally in IID, and 0.98 to 1.02 times locally and 1.00 to 1.20 times globally in non-IID settings. With width variations, \textit{FedFA exceeds HeteroFL}, achieving 1.01 to 1.16 times better accuracy globally in IID, 0.99 to 1.13 times locally, and 1.00 to 1.16 times globally in non-IID settings. For combined width and depth changes, \textit{FedFA surpasses NeFL}, showing a 1.00 to 1.02 times improvement globally in IID, 0.99 to 1.06 times locally, and 0.95 to 1.06 times globally in non-IID settings. Overall, \textit{FedFA outperforms other heterogeneous strategies in testing accuracy} except in 2 out of 9 scenarios.

As shown in Figure~\ref{figure: graph teest results}, backdoor attack scenarios reveal more distinct differences. When varying only the depth, FlexiFed experiences a more significant drop in testing accuracy than FedFA, with decreases of 1.22 to 2.09 times globally in IID, 2.21 to 3.31 times locally, and 1.21 to 1.74 times globally in non-IID settings. With width variation only, HeteroFL sees a testing accuracy drop of 1.07 to 1.47 times globally in IID, 0.89 to 2.37 times locally, and 1.11 to 1.22 times globally in non-IID settings compared to FedFA.  When the accuracy was 0.89 times higher (specifically in the CIFAR-100 local test scenario), FedFA consistently achieved much greater accuracy than HeteroFL under normal and severe attack conditions. For scenarios with width and depth variation, NeFL shows a drop in testing accuracy of 1.01 to 2.11 times globally in IID, 1.30 to 2.65 times locally, and 1.23 to 1.63 times globally in non-IID settings compared to FedFA. To summarize, FedFA generally surpasses other heterogeneous strategies in testing accuracy except in 3 out of 27 scenarios. These findings indicate that \textit{FedFA is remarkably robust against backdoor attacks on the global model}. 

Lastly, to demonstrate the generality of FedFA, we also examine its performance with transformers. The earlier strategies exhibit perplexities that are 1.07 to 4.50 times higher than those of FedFA, as detailed in Table~\ref{table: test results transformer}.

\begin{table}[t]
    \centering
    \caption{Average local testing perplexities for Transformer on the WikiText-2 dataset. FedFA's three variants yield much lower perplexities than its competitors.}
    \includegraphics[width=\linewidth]{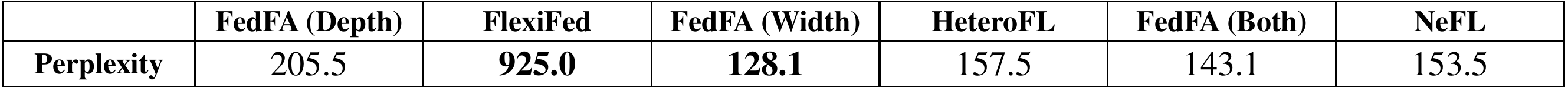}
    \label{table: test results transformer}
\end{table}

\subsubsection{Computational Complexity}
FedFA, employing layer grafting and scalable aggregation, has slightly higher computational complexity than earlier heterogeneous methods. Yet, for targeted testing accuracies—70\% (IID) and 40\% (non-IID) in CIFAR-10, 25\% (both IID and non-IID) in CIFAR-100, and 80\% (IID) and 30\% (non-IID) in Fashion MNIST—the computational complexities are only 0.95 to 1.02 times higher. This indicates that FedFA's computational overhead is not marginally higher than earlier strategies, as presented in Table~\ref{table: computation results}.

\section{Future Work and Limitations}\label{sec:future}
Future research could enhance FedFA's scalability for larger and more complex networks by optimizing algorithms to reduce the communication overhead and computational burden on clients. Developing dynamic client participation algorithms based on resource availability and network conditions could improve resource efficiency and model convergence speed but require careful mechanisms to handle clients' heterogeneous architectures. Advanced security mechanisms are necessary to detect and mitigate a broader range of adversarial attacks beyond the backdoor attacks we consider, including those exploiting model aggregation vulnerabilities. Further personalizing model architectures based on client data characteristics using advanced NAS techniques could improve local model performance. Additionally, integrating FedFA with edge computing paradigms could address latency, bandwidth, and real-time processing challenges in highly distributed environments.

Despite its advantages, FedFA has limitations that need to be addressed. One significant limitation is that all clients must employ the same type of network architecture, such as all using ResNets, MobileNets, or EfficientNets. This uniformity can restrict the flexibility and efficiency of the system, especially when dealing with diverse client capabilities and requirements. Future work should focus on enabling support for heterogeneous network types within the same federated learning framework to accommodate the variety of client devices better and improve overall performance and scalability.

\section{Conclusion}\label{sec:conclusion}
This paper introduces FedFA, a width- and depth-flexible aggregation strategy designed for clients with diverse computational and communication requirements and their local data. FedFA safeguards the global model against backdoor attacks through the layer grafting method. Furthermore, it introduces a scalable aggregation method to manage scale variations among networks of differing complexities. Compared to previous heterogeneous network aggregation methods, FedFA has shown superior testing performance and robustness to backdoor attacks, establishing its feasibility as a solution for various FL applications.

\section*{Acknowledgements}
We thank A. Datta and P. Mardziel for access to computing resources for completing our experiments. This work was partially supported by the National Science Foundation under grants CNS-1751075, CNS-2106891, and CNS-2312761.
\bibliographystyle{splncs04}
\bibliography{mybibliography}

\appendix
\onecolumn
\newpage
\clearpage
\section{Testing Specifications}
\begin{table*}[ht]
    \caption{Details of the network architectures for Pre-ResNet, MobileNetV2, EfficientNetV2, and Transformer are presented.}
    \centering
    \includegraphics[width=10 cm]{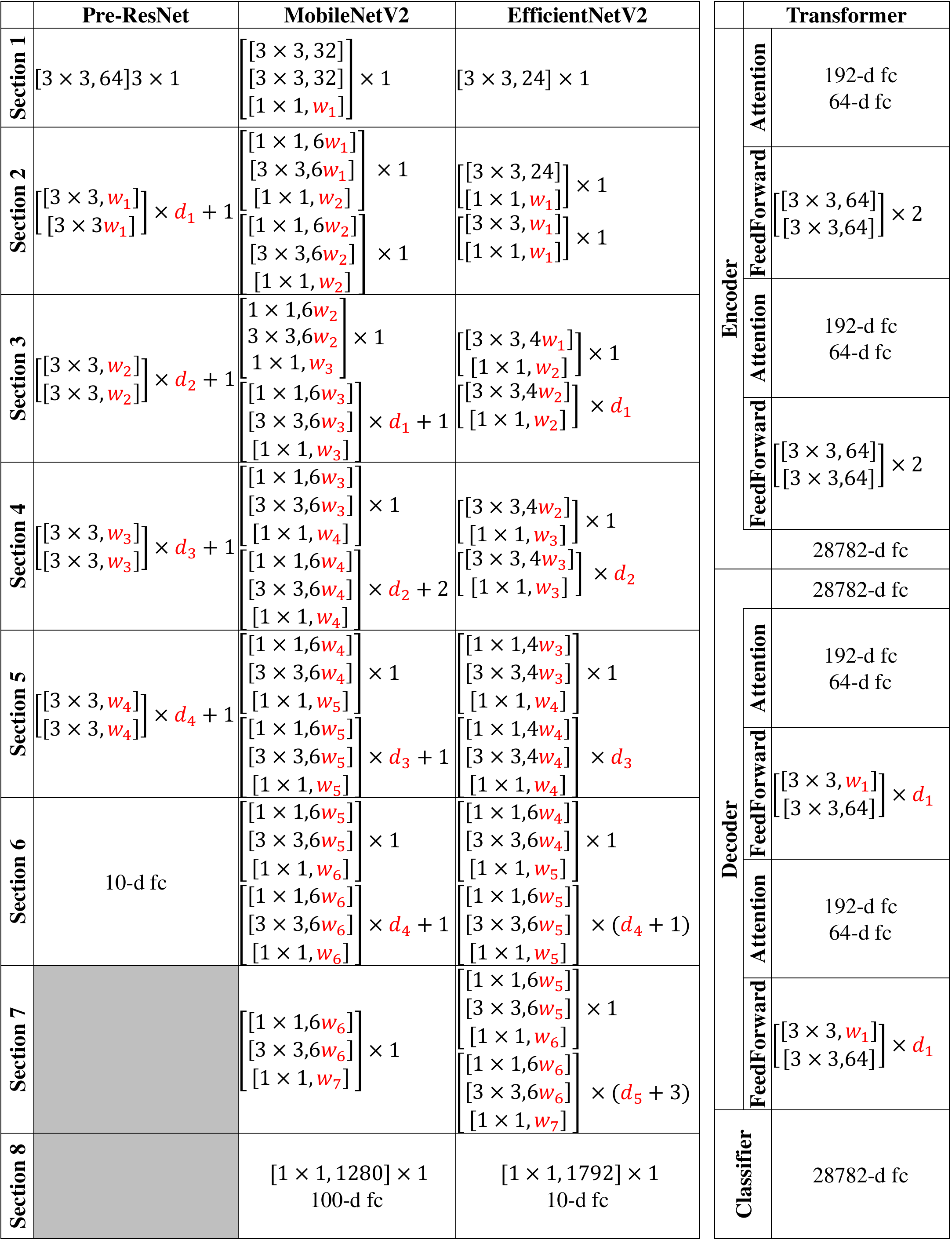}
    \label{figure: model archi 1}
\end{table*}

\begin{table*}[ht]
    \caption{The candidate values for the networks' depth $d_k$ and width $w_k$ in Table~\ref{figure: model archi 1}.}
    \centering
    \includegraphics[width=\linewidth]{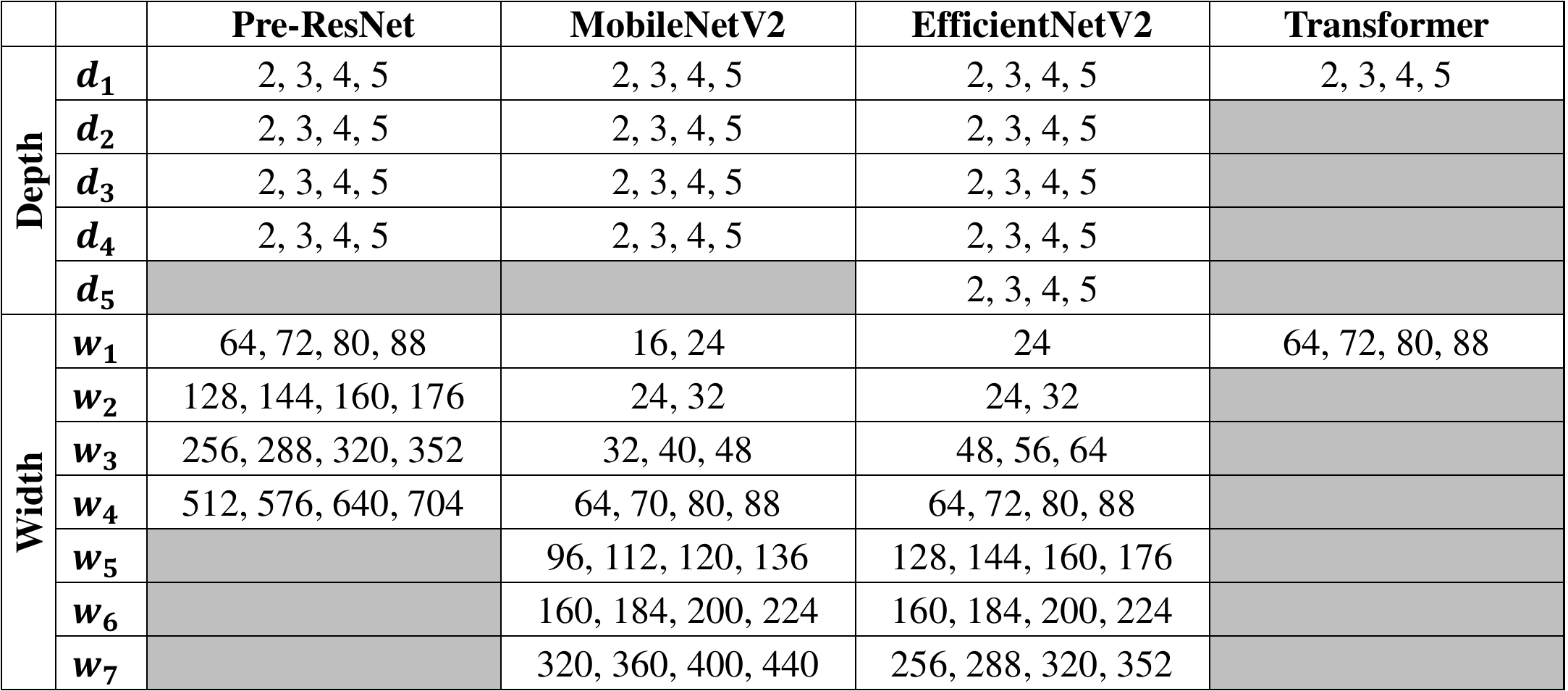}
    \label{figure: model archi 2}
\end{table*}

\begin{table*}[!ht]
    \centering
    \caption{Test conditions for evaluating layer similarity in Section~\ref{apdx: layer grafting}, scale variations in Section~\ref{apdx: scale variation} in the Appendix, and performance comparisons in Section~\ref{sec:experimental evaluation}.}
    \includegraphics[width=\linewidth]{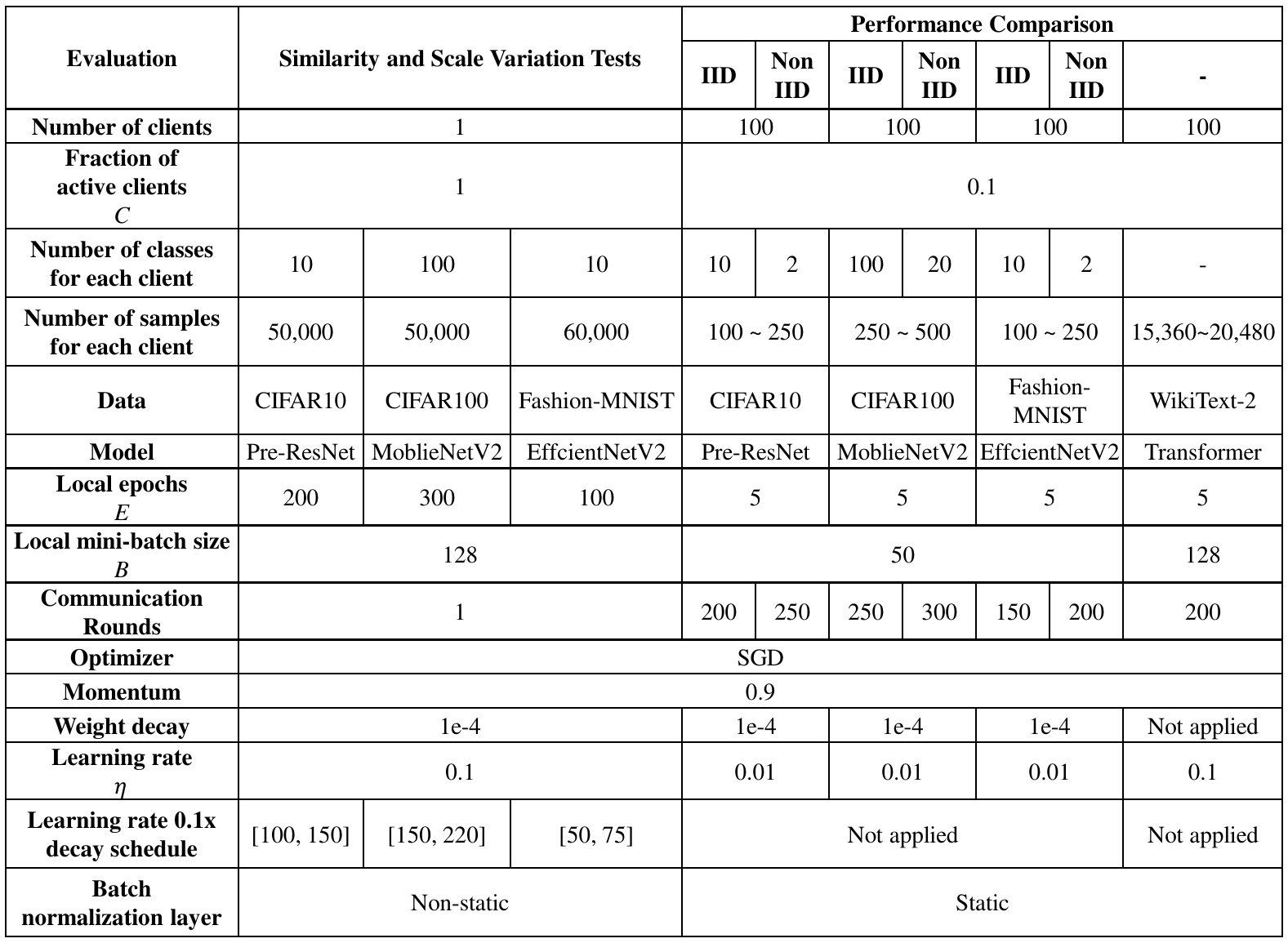}
    \label{figure: test condition}
\end{table*}
\newpage
\clearpage
\section{Understanding Residual Blocks and Skip Connections in Convolutional Neural Networks} 
\label{apdx: layer grafting}
\subsection{Structure and Properties of Convolutional Neural Networks with Skip Connections}
Typical Convolutional Neural Network (CNN) architectures comprise a sequence of input layers, followed by sections of residual blocks, and concluding with output linear layers. A key feature in many such architectures is the use of skip connections, which connect layers across these residual blocks.
\begin{figure}[ht]
    \centering
    \includegraphics[width=\linewidth]{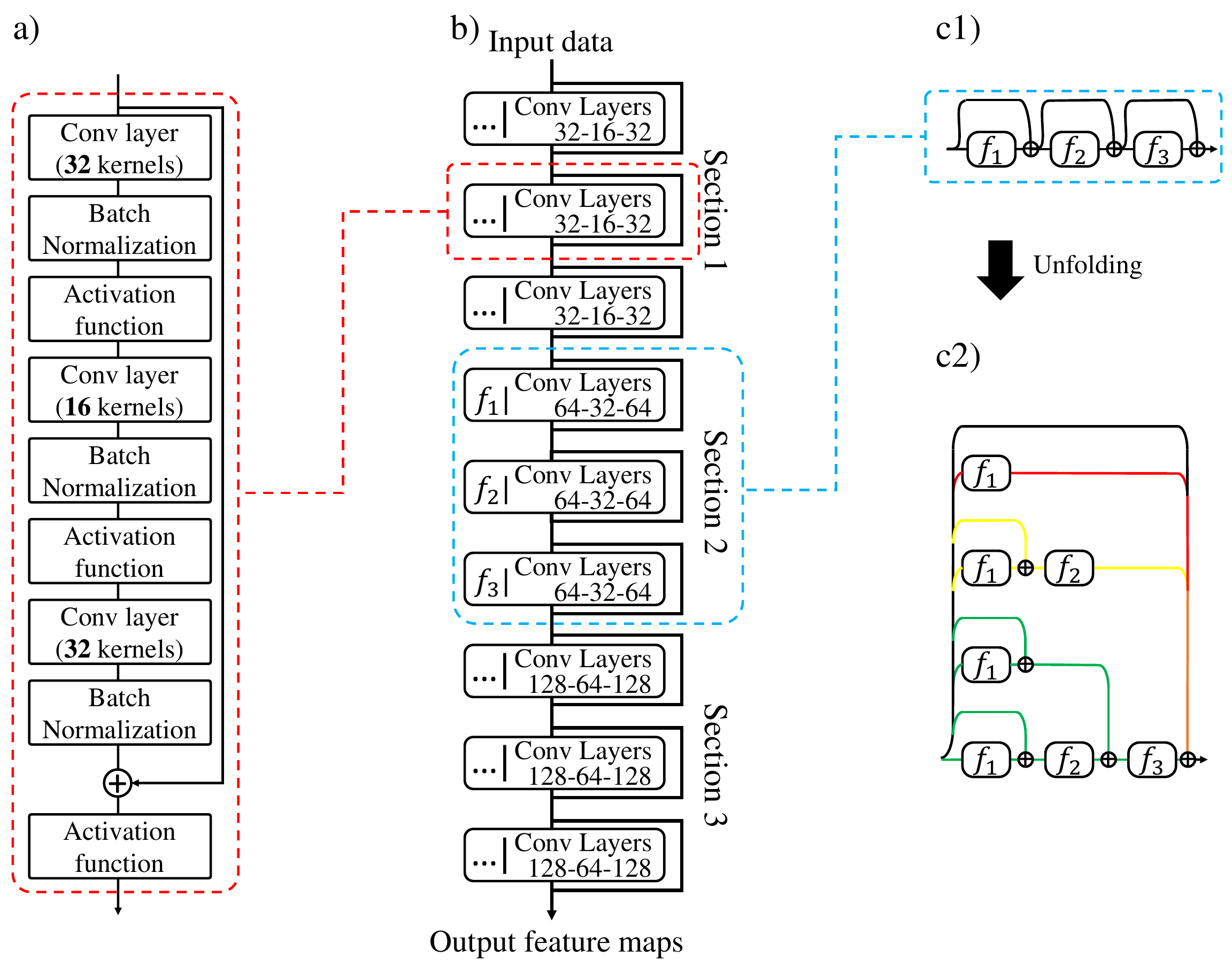}
    \caption{a) A residual block in a CNN, including convolutional and batch normalization layers. b) A skip connection network with residual blocks. c1) Residual blocks from Section 2 of the network. c2) The unfolded network, highlighting how skip connections enable an ensemble-like system.}
    \label{fig: skip connection ensemble}
\end{figure}
The pivotal property of CNNs with skip connections, and the focus of our layer grafting approach, is that CNNs with skip connections exhibit residual blocks with similar weight values. This similarity makes it possible to graft the last residual blocks of each section, playing a crucial role in guiding the aggregation of local models of different depths and informing the selection of layers during global model dissemination to local clients.

\subsection{Residual Block Similarity and Its Implications}
The similarity among residual blocks within a given section of a CNN can be empirically observed and has significant implications for model performance. Veit et al.~\cite{veit2016residual}'s results suggest that CNNs' performance is not drastically affected by removing or swapping certain residual blocks. This observation is crucial in understanding the resilience of CNNs with skip connections and forms the basis of our layer grafting strategy.

Consider a skip connection network as shown in Figure~\ref{fig: skip connection ensemble} b), divided into sections with their residual blocks ($f_x$, for $x = 1,2,3$). Each block within a section typically employs a similar convolutional layer structure, and the model's output can be conceptualized as an average of results from various sub-model paths formed by these blocks.

For an input $x$, the output of a section with residual blocks can be expressed as:
\begin{equation*}
    \begin{aligned}
        output = x + f_1(x) &+ f_2(x) + f_3(x) + f_2(f_1(x)) \\
        &+ f_3(f_1(x)) + f_3(f_2(x)) + f_3(f_2(f_1(x)))
    \end{aligned}
\end{equation*}
Swapping blocks, say $f_1$ and $f_2$, alters the equation but does not significantly affect the output, suggesting that $f_1(x)$ and $f_2(x)$ have similar contributions. This observation leads us to an important relationship:
\[
    f_2(f_1(x)) + f_3(f_2(f_1(x))) \approx f_1(f_2(x)) + f_3(f_1(f_2(x)))
\]
Algebraic manipulation then brings us to the conclusion that $f_1\approx f_2\approx f_3$: \textit{residual blocks within a given CNN are similar}. This reasoning can be generalized to larger numbers of residual blocks with various layer structures.

\subsection{Statistical Evidence for Layer Similarity}
We also statistically validate layer similarity. We consider the three different CNNs shown in Table~\ref{figure: model archi 1}: Pre-ResNet, MobileNetV2, and EfficientNetV2, which are respectively trained on the CIFAR-10, CIFAR-100, and Fashion MNIST datasets. We use models of different depths $d_k$, with $d_k$ determining the number of residual blocks in each convolutional layer section, as shown in the table.

We do this by using the Pearson Correlation Coefficient (PCC). While the PCC has been used to assess correlations between filter weights, e.g., to facilitate model pruning~\cite{kumar2022corrnet}, we directly utilize the PCC to gauge the similarity between convolutional layers and thus residual blocks. A high PCC value, combined with our previous result on similarities in weight scales, would indicate similarities in residual block weights.

To measure the similarity of two residual blocks, we first consider using the PCC to assess the similarities of two convolutional layers, $\mathbb{A}$ and $\mathbb{B}$, in distinct residual blocks. Each layer contains $N_{Cout}$ filters (i.e., output channels), with each filter possessing $N_{Cin}$ weight maps (i.e., input channels). We use $r^{i,j}_{k,l}$ to represent the PCC value of the $k$-th weight map of the $i$-th filter of layer $\mathbb{A}$ and the $l$-th weight map of the $j$-th filter of layer $\mathbb{B}$.

We use $\mathbf{R}^{i,j}$ to denote the matrix of the PCCs for all pairs of weight maps in the $i$-th filter of layer $\mathbb{A}$ and the $j$-th filter of layer $\mathbb{B}$. Our goal is now to represent $\mathbf{R}^{i,j}$ with a single scalar encapsulating the overall similarities of filters $i$ and $j$.

\begin{figure}[ht]
    \centering
    \includegraphics[width=9 cm]{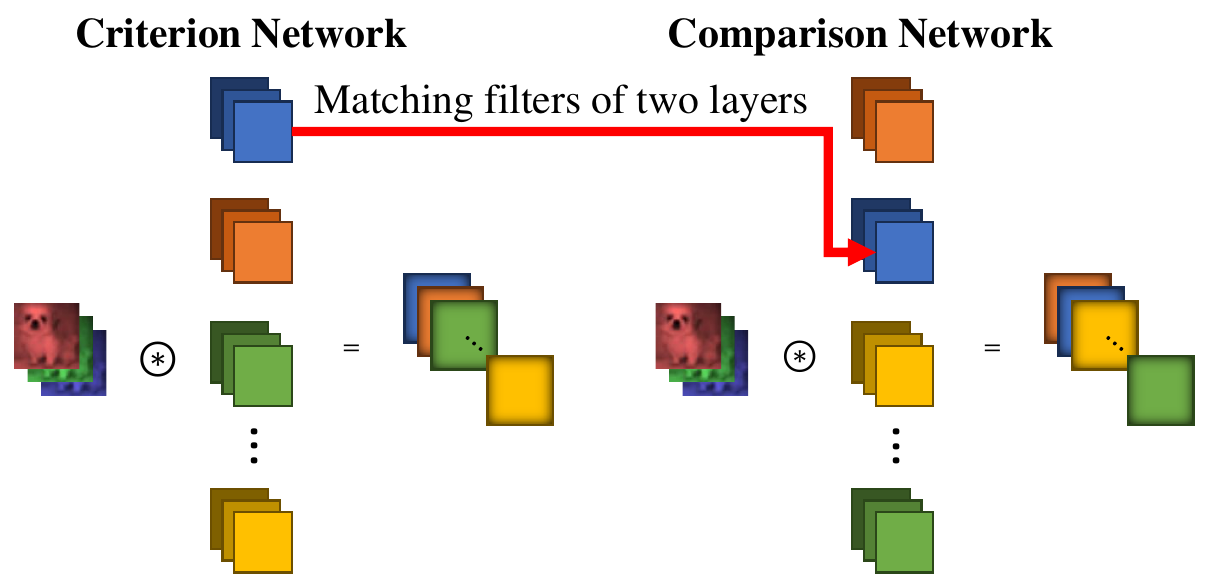}
    \caption{Randomness in sequences of filters and weight maps in convolutional layers.}
    \label{fig: similarity cases}
\end{figure}

Since weights in networks are initialized randomly for every new training iteration, filter sequences can differ, even with identical input feature maps, leading to diverse output feature map sequences (Figure~\ref{fig: similarity cases})~\cite{glorot2010understanding,he2015delving}. Importantly, these outputs then serve as input for subsequent convolutional layers, influencing the sequence of weight maps within each filter.

To match these sequences, we thus select the element with the highest PCC in each row of $\mathbf{R}^{i,j}$, with the constraint that only one element per column of $\mathbf{R}^{i,j}$ is selected. We then compute the average of the selected $N_{Cin}$ elements, denoted as $\hat{r}^{ij}$. Since we need to account for potential overlapping weight maps, any column with a selected element is excluded in subsequent steps.

When assessing two convolutional layers with $N_{Cout}$ filters each, there are $N_{Cout}^2$ filter pair combinations. Just as with weight map similarity, we create a one-to-one filter matching, since previous research has highlighted the existence of many similar filters within a single layer~\cite{kumar2022corrnet}.
This one-to-one matching avoids matching a given filter with an excessive number of overlapping filters and overstating the overall layer similarity. We then average the PCCs $\hat{r}^{i,j}$ of the matched filters.

The similarities between the convolutional layers of two models with varying depths are presented in Tables~\ref{fig: similarity table 1}, \ref{fig: similarity table 2}, and \ref{fig: similarity table 3}. We consider models at the 0 epoch (i.e., before training) and at the epoch where they achieve their highest testing accuracy. Specifically, we exclude the first residual block of each section from our analysis, as it typically has a different input channel size compared to the other residual blocks in the same section. 

From Tables~\ref{fig: similarity table 1}, \ref{fig: similarity table 2}, and \ref{fig: similarity table 3}, all convolutional layers within a particular section have similarities greater than 0.5, regardless of the presence of skip connections. CNNs with skip connections usually show increased correlations post-training, with exceptions in 43 of the 138 cases (emphasized in bold). In contrast, CNNs without skip connections reveal decreased correlations post-training in 76 of the 138 cases (emphasized in bold).

Prior to training, high similarities are observed in both types of CNNs, suggesting potential matching of filters or weight maps, possibly due to the law of large numbers. This trend is even more evident as filter counts rise from lower to higher sections. Notably, while CNNs with skip connections mostly show an increase in similarity, over half the cases without them deviate from this trend.

This data shows that \textit{the weights of residual blocks within the same section of CNNs with skip connections remain similar}, irrespective of depth or structural variations. We thus validate layer grafting to aggregate client models, as well as our method of sending model weight subsets to each client.

\subsection{Layer Grafting Based on Residual Block Similarity}
The observed similarity among residual blocks in CNNs allows us to infer that any block from the same section is a viable candidate for grafting. This similarity supports the layer grafting method in two ways:
\begin{itemize}
    \item \textbf{Depth Modification}: When modifying the depth of the global model for local models, any residual block from the same section can be used, ensuring consistency in feature representation and learning capability.
    \item \textbf{Model Adaptability}: The similarity in blocks enables a more flexible approach to model aggregation and dissemination, as blocks can be added or removed without significantly impacting model performance.
\end{itemize}

\subsection{Summary}
\begin{figure}[ht]
    \centering
    \includegraphics[width=\linewidth]{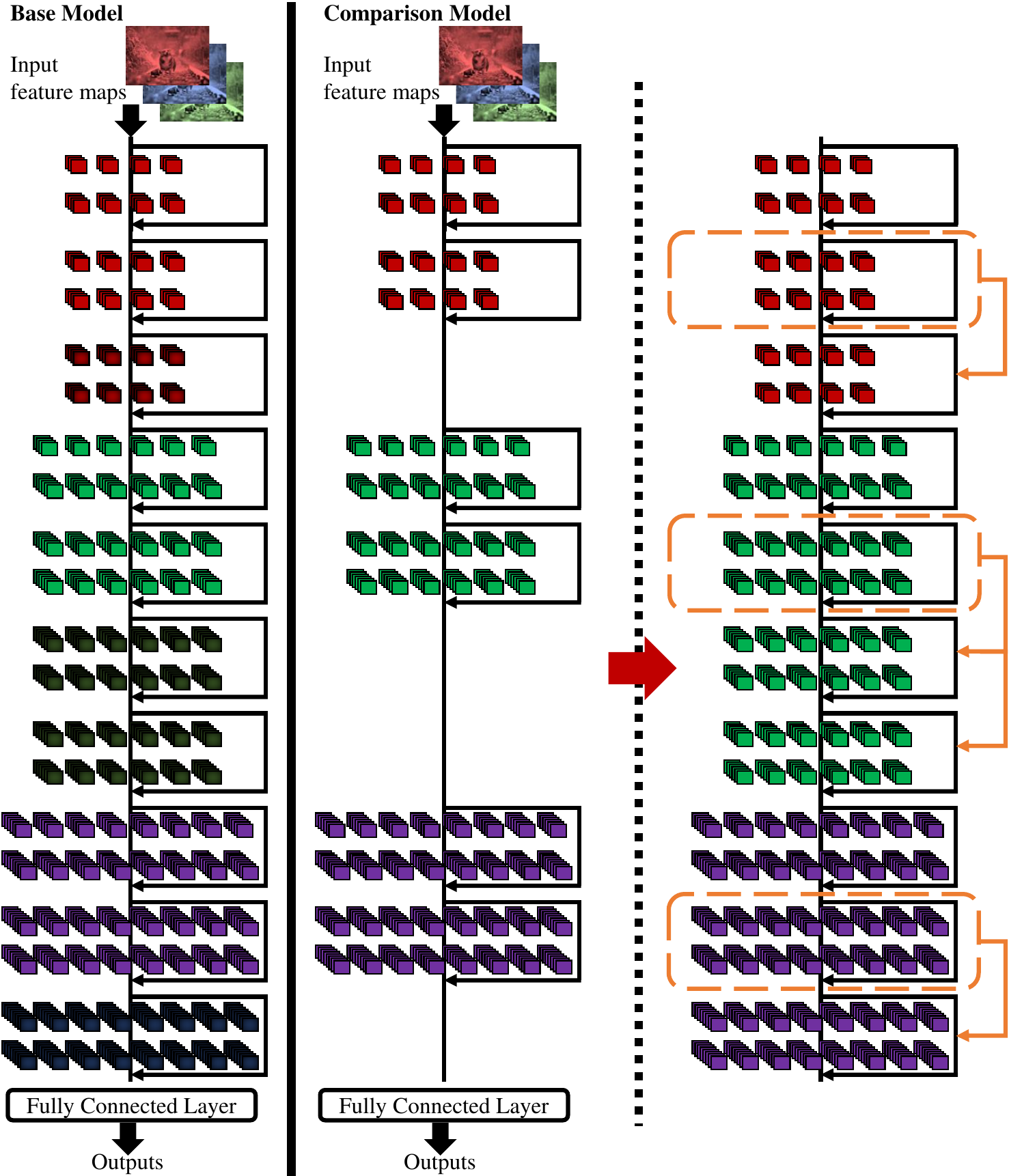}
    \caption{The layer grafting method to standardize model depth by identifying the maximum depths of sections and augmenting each client's model with additional residual blocks until this depth is reached.}
    \label{figure: layer grafting}
\end{figure}

The statistical analysis of the layer grafting method in CNNs with skip connections underscores the viability of this approach in FL. By leveraging the inherent similarity among residual blocks, the layer grafting method not only preserves model integrity but also enhances adaptability and robustness across diverse client models. This analysis affirms the soundness of layer grafting as a key component in our FedFA framework in Figure~\ref{figure: layer grafting}.

\begin{table}[!ht]
    \centering
    \caption{Similarities of convolutional layers for Pre-ResNet.}
    \includegraphics[width=8.3 cm]{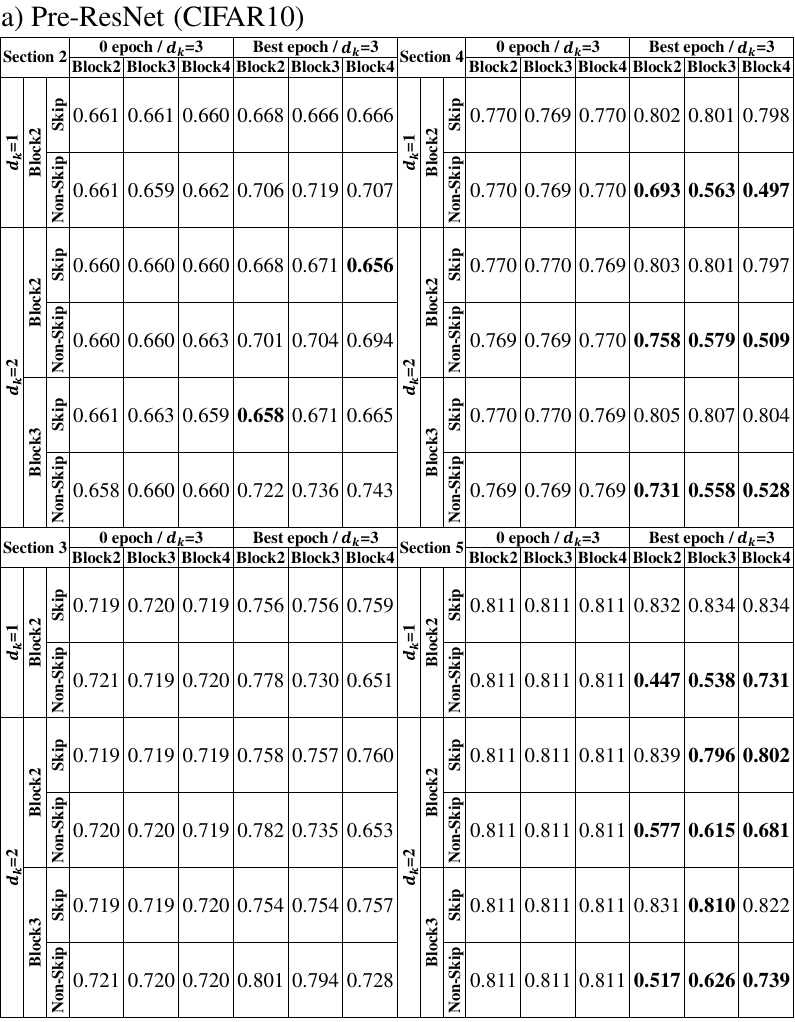}
    \label{fig: similarity table 1}
\end{table}
\begin{table}[!ht]
    \centering
    \caption{Similarities of convolutional layers for MobileNetV2.}
    \includegraphics[width=8.3 cm]{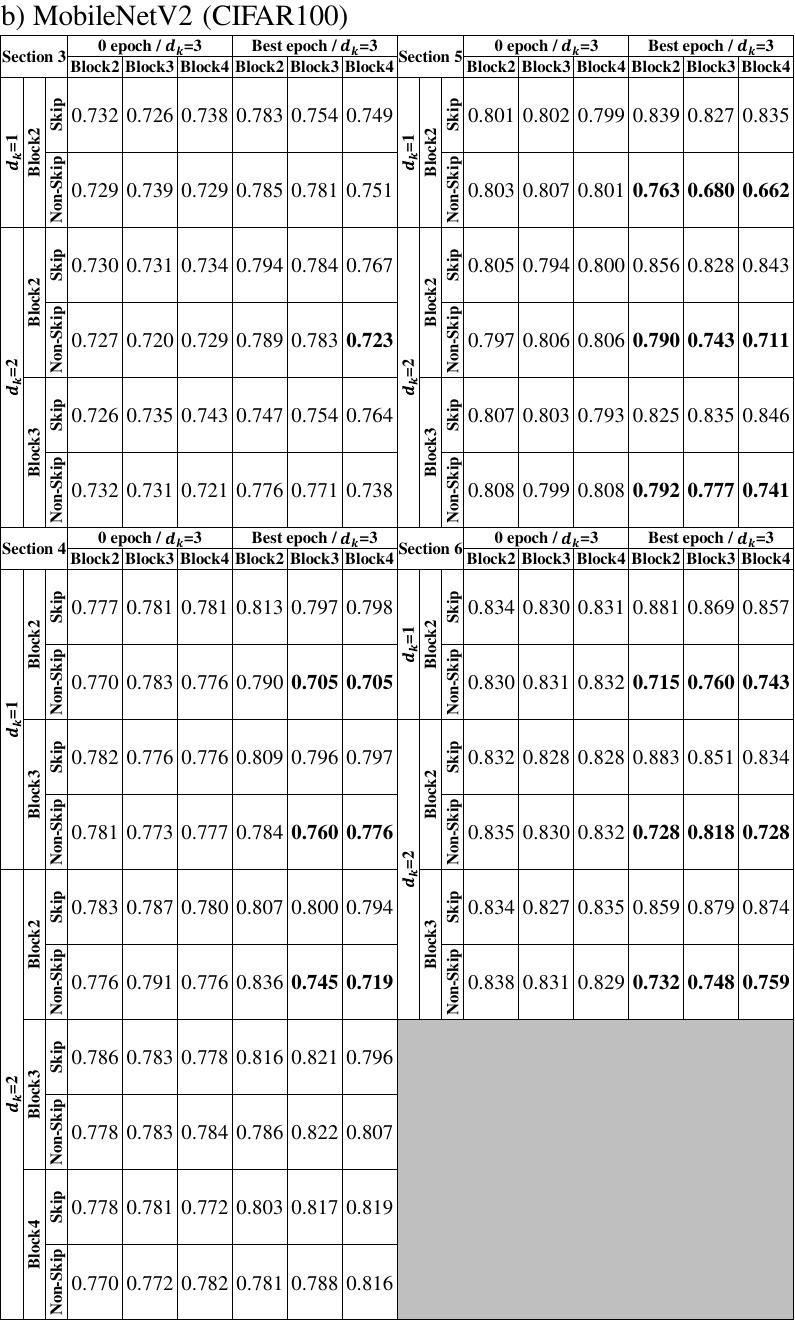}
    \label{fig: similarity table 2}
\end{table}
\begin{table}[!ht]
    \centering
    \caption{Similarities of convolutional layers for EfficientNetV2.}
    \includegraphics[width=8.3 cm]{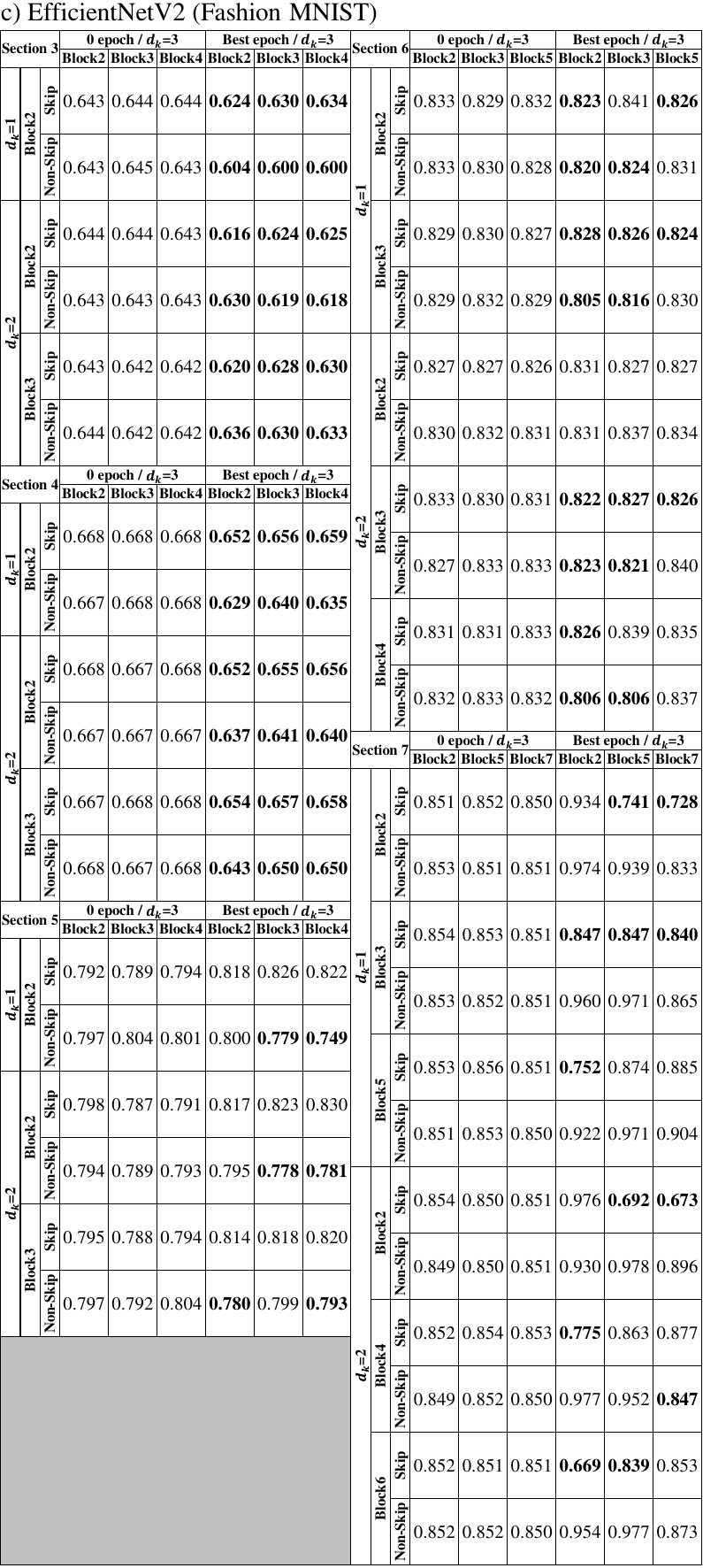}
    \label{fig: similarity table 3}
\end{table}

\clearpage
\newpage
\section{Global Model distribution Algorithm}
\begin{algorithm}[h]
\caption{This algorithm customizes the aggregated global model $M_G^t$ for each client $c$ by adjusting its structure to align with its original local network architecture, $M_c^{t-1}$ (line 7 of Algorithm~\ref{algorithm: FedFA}). The global model has the number of residual blocks for section $s$, $N_{depth,\max}^{(s)}$, whereas the local model, $M_c^{t-1}$, has the number of residual blocks for each section, $N_{depth,c}^{(s)}$ and the input and the output channel sizes for of each layer, $C_{I}$ and $C_{o}$. Customization involves reducing the depth and width of $M_G$ by systematically removing residual blocks and filters over the client-specific thresholds. In the algorithm, the $\ominus$ operator signifies the reverse of the layer grafting process.}
\label{algorithm: Model Distribution}
\begin{algorithmic}[1]
\REQUIRE Updated global model $M_G^t$.
\ENSURE Each client $c$ receives an appropriately configured version of $M_G^t$, $M_c^t$.
\FOR{each section $M_G^{(s)}$ in global model $M_G^t$}
    \STATE $\Delta D \gets N_{depth,\max}^{(s)} - N_{depth,c}^{(s)}$ 
    \FOR{$d = 1$ to $\Delta D$}
        \STATE $R_{last}^{(s)} \gets \text{last residual block in section } s$
        \STATE $M_G^{(s)} \gets M_G^{(s)} \ominus R_{last}^{(s)}$
    \ENDFOR
\ENDFOR
\FOR{each layer $M_G^{(l)}$ in global model $M_G^t$}
    \STATE$C_{I},C_{o}$ $\gets $ Input and output channel sizes of $M_c^{(l)}$ 
    \STATE $M_G^{(s)} \gets M_G^{(s)}[:C_{o},:C_{I}]$    
\ENDFOR
\STATE $M_c^t \gets M_G^t$ 
\end{algorithmic}
\end{algorithm}
\newpage
\clearpage
\section{Efficacy of Heterogeneous Network Aggregation} \label{apdx: heteroaggre effectiveness}
This section investigates the theoretical underpinnings of heterogeneous network aggregation. Due to their simpler structure, shallow networks demonstrate a faster convergence rate than deeper networks. However, deeper networks are more adept at capturing complex features and hierarchical data structures, translating to superior performance on complex tasks~\cite{mhaskar2016deep}. Therefore, aggregating shallow and deep models can accelerate convergence relative to only deep models and enhance performance compared to solely shallow models. This theoretical examination focuses on how the speed of increase in prediction variance (the output logits of the classifier) differs between shallow and deep models in classification tasks.

\subsection{Variance Analysis Across Model Complexities}
When a model is fully trained in classification tasks, its output logits typically approach $1$s for the correct classes and $0$s for others, assuming one-hot encoding. This indicates that the variance of the output logits generally increases until the models are completely adapted to their data. This subsection explores the relationship between model complexity and variance in model predictions for classification tasks, analyzing models of varying complexities to discern their impact on predictive variability.

\subsubsection{Preliminaries and Assumptions}
To guide our analysis, we define essential concepts and assumptions:
\begin{itemize}
    \item \textbf{Model Output Functionality}: In classification tasks, consider the model's output logits vector, \(y\), is a function of the input \(x\) and weight vectors \(w_i\) for the index of each layer $i$. We assume that the activation function $g$ negates the correlated outputs of \( w_i \) for analytical simplicity. Additionally, appropriately clipped weights \( w_i \) can compute the output \( y \) without needing a softmax function. Thus, \( y \) can be expressed as a linear combination of \( g(w_i, x) \), formulated as \( y = \sum_i g(w_i, x) \). Also, the analysis assumes the optimization process employs full batch gradient descent.
    \item \textbf{Law of Large Numbers (LLN) Application}: The LLN indicates that as the number of trials increases, the average of the outcomes converges to the expected value. This principle is instrumental in understanding the variance behavior with increasing model complexity.
\end{itemize}

\subsubsection{Model Complexity and Variance}
We explore the relationship between model complexity (number of weights) and variance in a structured manner. The total variance of \(y\) is articulated through the law of total variance:
\[ \text{Var}(y) = E[\text{Var}(y|x)] + \text{Var}(E[y|x]) \]
Here, \(E[\text{Var}(y|x)]\) signifies the expected value of the conditional variance of \(y\) given \(x\), and \(\text{Var}(E[y|x])\) represents the variance of the expected value of \(y\) given \(x\).

Analyzing each component, assuming \(g(w_i, x)\) is independent for each \(i\):\\
1. Expectation of Conditional Variance \(E[\text{Var}(y|x)]\)
\begin{align*}
 E[\text{Var}(y|x)] &= E[\text{Var}(\sum_i g(w_i, x)|x)] \\
 &= E[\sum_i \text{Var}(g(w_i, x)|x)] + 2E[\sum_{i\neq j} \text{Cov}(g(w_i, x), g(w_j, x)|x)] \\
 &= E[\sum_i \text{Var}(g(w_i, x)|x)] +0 = \sum_i E[\text{Var}(g(w_i, x)|x)]   
\end{align*}
2. Variance of the Expected Value \( \text{Var}(E[y|x]) \)
\[ \text{Var}(E[y|x]) = \text{Var} \left( E[\sum_i g(w_i,x)|x] \right) = \text{Var} \left(\sum_i E[ g(w_i,x)|x] \right) \]
Given that \(g(w_i,x)\) is independent for each \(i\):
\[ \text{Var}(E[y|x]) = \sum_i \text{Var} \left( E[g(w_i,x)|x] \right) \]
Combining both components yields:
\[ \text{Var}(y) = \sum_i E[\text{Var}(g(w_i, x)|x)] + \sum_i \text{Var} \left( E[g(w_i,x)|x] \right) \]

\subsubsection{Averaging Effect of Weights}
The LLN posits that as the number of weights \(n\) increases, the individual contributions to the output variance by each weight, \(\text{Var}(g(w_i,x)|x)\), average out. This results in a reduction of the overall variance attributable to any single weight.

\subsubsection{Comparative Analysis of Deep and Shallow Models}
Consider a deep model with a large number of weights (\(N_{\text{deep}}\)) and a shallow model with fewer weights (\(N_{\text{shallow}}\)). Assuming \(N_{\text{shallow}}\) is sufficient for the LLN to apply, we define the contribution of each weight to the output logits' variance for deep and shallow models as \(Var_{\text{deep}}\) and \(Var_{\text{shallow}}\), respectively:
\begin{align*}
\text{Var}_{\text{deep}}
&= \frac{1}{N_{\text{deep}}} \left( \sum_{i=1}^{N_{\text{deep}}} E[\text{Var}(g(w_i, x)|x)] + \sum_{i=1}^{N_{\text{deep}}} \text{Var}(E[g(w_i,x)|x]) \right) \\
&= \frac{1}{N_{\text{deep}}}\text{Var}(y)
\end{align*}
\begin{align*}
\text{Var}_{\text{shallow}} 
&= \frac{1}{N_{\text{shallow}}} \left( \sum_{i=1}^{N_{\text{shallow}}} E[\text{Var}(g(w_i, x)|x)] + \sum_{i=1}^{N_{\text{shallow}}} \text{Var}(E[g(w_i,x)|x]) \right) \\
&= \frac{1}{N_{\text{shallow}}}\text{Var}(y)
\end{align*}
For a given target variance \(\text{Var}(y)\), and considering \(N_{\text{deep}} > N_{\text{shallow}}\), the LLN implies that the average variance attributable to individual weights in the deep model is less than that in the shallow model:
\[
 \frac{1}{N_{\text{deep}}}\text{Var}(y) \le \frac{1}{N_{\text{shallow}}}\text{Var}(y)
\]
Consequently:
\[
\text{Var}_{\text{deep}} \le \text{Var}_{\text{shallow}}
\]

\subsection{Post-Training Variance Dynamics}
This section delves into the increase in the variance of output logits post-training, reflecting a shift from initial model outputs to a more diversified set of predictions due to learning from the training data.

\subsubsection{Foundational Premises}
\begin{itemize}
    \item \textbf{Weight Initialization}: Before training, weights \(w_i\) are uniformly initialized (e.g., all ones), leading to a deterministic output for any given input \(x\), assuming \(w_i\) remains fixed. Nevertheless, in a more complex model where the output nonlinearly depends on \(w_i\), even with uniform initialization, there exists a potential for non-zero variance in the output if the model's function \(\sum_i g(w_i, x)\) maps the same \(w_i\) to varying outputs for different inputs \(x\).
    \item \textbf{Training Dynamics}: During training, \(w_i\) is adjusted to minimize a loss function \(L(y, y^*)\), where \(y^*\) represents the target output vector. This adjustment adheres to the full batch gradient descent methodology at iteration $t$, represented as:
    \[ w_{i,t+1} = w_{i,t} + \eta \nabla L (y_t , y^*) \]
    Here, \(\eta\) is the learning rate.
\end{itemize}

\subsubsection{Pre-Training Variance Dynamics}
Before training commences, all weights \(w_i\) (i.e., $w_{i,0}$) are initialized to a uniform value, leading to a theoretical scenario where the model's output variance might show minimal but non-zero values due to the differential mapping of inputs \(x\) by \(\sum_i g(w_{i,0}, x)\). The initial assumption is that:
\[ \text{Var}(y_0) = \text{Var}(\sum_i g(w_{i,0}, x)) \approx 0 \]
In a simple or linear model where \(f\) represents the model function, \(\text{Var}(y_0)\) could be nearly zero if \(\sum_i g(w_{i,0}, x)\) yields consistent outputs. However, this variance may not be negligible for more complex models, though it remains relatively small.

\subsubsection{Post-Training Variance Dynamics}
After training, the weights \(w_i\) are updated to minimize the loss function \(L\), introducing variability in \(w_{i,t+1}\) which, in turn, injects diversity into the model's output:
\[ w_{i,t+1} = w_{i,t} + \eta \nabla L (y_t, y^*) \quad \text{and} \quad L (y_{t+1}, y^*) \le L (y_t, y^*) \]
As \(w_{i,t+1}\) is continually adjusted to reduce \(L\), reflecting the learning process from the training data, the variance \(\text{Var}(y_{t+1})\) tends to increase, acknowledging the diversity in responses due to this learned variability:
\[ \text{Var}(y_{t+1}) \ge  \text{Var}(y_{t}) \]
for the training data \(x\).

\subsection{Toward a Specific Target Variance}
This subsection systematically examines how the weights of deep and shallow models affect the speed toward a specific target variance \(\text{Var}_{\text{target}}\) and finally shows that aggregating deep and shallow models is beneficial.

From earlier discussions:
\[ \frac{1}{N_{\text{shallow}}} \text{Var}(y_t) = \text{Var}_{\text{shallow},t} \ge \text{Var}_{\text{deep},t} = \frac{1}{N_{\text{deep}}} \text{Var}(y_t) \]
\[ \text{Var}(y_{t+1}) \ge \text{Var}(y_{t}) \]
We know from the given conditions that the accumulated variances for shallow and deep models equalize at certain times \( T_{\text{shallow}} \) and \( T_{\text{deep}} \) for $y^*$, respectively:
\[ \text{Var}(y^*) =  N_{\text{shallow}}  \text{Var}_{\text{shallow},T_{\text{shallow}}} = N_{\text{deep}}  \text{Var}_{\text{deep},T_{\text{deep}}} \]
Let \( r_{\text{shallow}} \) and \( r_{\text{deep}} \) represent the average rates of variance growth for shallow and deep models, respectively. These rates are defined as the change in variance per training iteration. Over time, the variances for shallow and deep models can be represented as a function of their growth rates and time:
\[ \text{Var}_{\text{shallow},T_{\text{shallow}}} = r_{\text{shallow}}  T_{\text{shallow}} \]
\[ \text{Var}_{\text{deep},T_{\text{deep}}} = r_{\text{deep}}  T_{\text{deep}} \]
Plugging these into the equation of accumulated variances gives:
\[ N_{\text{shallow}}  r_{\text{shallow}}  T_{\text{shallow}} = N_{\text{deep}}  r_{\text{deep}}  T_{\text{deep}} \]
To find a direct relationship between \( r_{\text{shallow}} \) and \( r_{\text{deep}} \), rearrange the equation:
\[ r_{\text{shallow}} = \frac{N_{\text{deep}}  T_{\text{deep}}}{N_{\text{shallow}}  T_{\text{shallow}}}  r_{\text{deep}}\]
and knowing \( T_{\text{deep}} \ge T_{\text{shallow}} \)~\cite{mhaskar2016deep}, Therefore, we can infer that:
\[ r_{\text{shallow}} \ge r_{\text{deep}} \]
This implies that shallow models exhibit a faster increase in variance compared to deep models, illustrating that the variance in predictions of a shallow model increases more quickly during training.

\subsubsection{Aggregation Dynamics of Deep and Shallow Model Weights}
The aggregation of weights from both deep and shallow models creates a new dynamic in the variance increase rate of the combined model. This can be quantified as:
\[
r_{\text{combined}} = \zeta  r_{\text{deep}} + (1 - \zeta) r_{\text{shallow}} \ge r_{\text{deep}}
\]
where \(\zeta\) is a factor that balances the contributions of deep and shallow model weights with \(0 < \zeta < 1\). This formula shows that combining the weights of deep and shallow models in the combined model configuration yields a more marked increase in the variance rate per weight compared to using only the deep model's weights.

\clearpage
\newpage
\section{Convergence Analysis of Gradient Aggregation in Federated Learning with Flexible Architectures} \label{apdx: convergence analysis}
In this section, we show the convergence rate of our FedFA's object function in the context of FL~\cite{mcmahan2017communication}. 
\subsubsection{Preliminary Concepts}
\begin{enumerate}
    \item \textbf{Mitigation of Skewness Introduced by Data Heterogeneity}:
   Different client local weights and data distributions, denoted as $ \omega_1, \omega_2, \dots, \omega_c$ and $ D_1, D_2, \dots, D_c $, yield gradients $ \nabla f_1, \nabla f_2, \dots, \nabla f_c $ that can vary significantly in magnitude. Formally:
    \[ \nabla f_c(\omega_c, D_c) = \frac{\partial}{\partial \omega_c} f_c(\omega_c, D_c) \]
    To counteract this variability, we employ the L2 norm for scaling. The L2 norm is calculated based on the central 95\% of the weights under the 95th percentile:
    \[ \nabla f_c^{\mathrm{scaled}} = 
    \left( \frac{\frac{1}{m} \sum_{\kappa \in \mathcal{C}_{sel}} \left\| \omega_{95\%, \kappa} \right\|}{\left\| \omega_{95\%, c} \right\|} \right) \nabla f_c(\omega_c, D_c) 
    = \alpha_c\nabla f_c(\omega_c, D_c)
    \]
    Unlike the FedFA's complete Algorithm~\ref{algorithm: FedFA}, we assume $\omega_c$ is a one-layered neural network architecture for simplicity. Let $|\mathcal{C}|$ symbolize the total client count within the FL framework. $C$ is the participating rate of clients for each round. $\mathcal{C}_{sel}$ indicates the client indices participating in a given round, and $m$ is the count of these clients, such that $m = |{C}_{sel}| $. It is ensured that $ m = C\times|\mathcal{C}| $.
    \item \textbf{Preservation of Gradient Direction}:    
    After scaling, the gradient's direction remains unchanged:
    \[ \text{Direction}(\nabla f_c^{\text{scaled}}) = \text{Direction}(\nabla f_c) \]
    This conservation guarantees that the unique insights from each client's local data are maintained post-scaling.
    \item \textbf{Robustness to Outliers}:
    Utilizing the central weights under the 95th percentile for scaling ensures resilience against extreme gradient values:
    \[ \omega_{95\%, c} = \text{Central 95\% of the weights} \]
    This mechanism normalizes potential outliers, preventing them from disproportionately influencing the global update.
    
    \item \textbf{Global Model Stability}:
    By aggregating the scaled gradients across all clients, we derive the global gradient as:
    \[ \nabla f_{G,\text{w/ scaled}} = \sum_{c\in{C}_{sel}} p_c \times \nabla f_c^{\text{scaled}} \]
    Here, $ p_c $ denotes weights, typically reflecting the data distribution of client $ c $.
\end{enumerate}

\subsubsection{Convergence Analysis}
To understand the convergence of our algorithm in federated learning, we analyze it under two primary mathematical properties: strong convexity and Lipschitz continuity of the gradient. \\

\noindent \textbf{Assumptions} 
\begin{enumerate}
  \item The global loss function $ f(\omega) $ is $ \mu $-strongly convex, where $ \mu $ is a positive real number, $\omega$ is weights. This property guarantees a unique minimum for $ f(\omega) $, which aids in the convergence of the optimization process.
  \item The gradient of $ f(\omega) $ exhibits Lipschitz continuity with a constant denoted as $ L $. Given that $ L $ is a non-negative real number, this continuity ensures that the gradient variations between consecutive iterations are bounded, ensuring stability during the optimization updates.
\end{enumerate}
\textbf{Analysis}
If a function $ f(\omega) $ is $ \mu $-strongly convex, then for every $ \omega_1 $ and $ \omega_2 $:
\[
f(\omega_1) \geq f(\omega_2) + \langle \nabla f(\omega_2), \omega_1 - \omega_2 \rangle + \frac{\mu}{2} \lVert \omega_1 - \omega_2 \rVert^2 
\]
The function of $ f(\omega) $ is $ L $-Lipschitz continuous when:
\[
\lVert f(\omega_1) - f(\omega_2) \rVert \leq L \lVert \omega_1 - \omega_2 \rVert 
\]
The iterative process of full batch gradient descent updates the weight $ \omega $ as:
\[
\omega_{t+1} = \omega_t + \eta \nabla f(\omega_t)
\]
Here, $ \eta $ is the learning rate. 
Thus, we can deduce:
\[ \lVert \omega_{t+1} - \omega_t \rVert = \eta \lVert \nabla f(\omega_t) \rVert \]
We start with the strong convexity property:
\begin{align*}
f(\omega_{t+1}) &\geq f(\omega_t) + \langle \nabla f(\omega_t), \eta \nabla f(\omega_t) \rangle + \frac{\mu}{2} \lVert \eta \nabla f(\omega_t) \rVert^2 \\
&= f(\omega_t) + \eta \lVert \nabla f(\omega_t) \rVert^2 + \frac{\mu \eta^2}{2} \lVert \nabla f(\omega_t) \rVert^2
\end{align*}

Considering the Lipschitz continuity, the function value change due to a step in the direction of the gradient is:
\begin{align*}
\lVert f(\omega_{t+1}) - f(\omega_t) \rVert \leq L \lVert \eta \nabla f(\omega_t) \rVert 
\end{align*}
Combining both inequalities, we get:
\begin{align*}
\lVert \eta \lVert \nabla f(\omega_t) \rVert^2 + \frac{\mu \eta^2}{2} \lVert \nabla f(\omega_t) \rVert^2 \rVert
&\leq \eta L\lVert \nabla f(\omega_t) \rVert  \\
\frac{2+\mu \eta}{2} \lVert \nabla f(\omega_t) \rVert &\leq L \\
\lVert \nabla f(\omega_t) \rVert &\leq \frac{2L}{2+ \mu \eta}
\end{align*}
This inequality gives us a bound on the magnitude of the gradient at iteration $ t $. If the magnitude of the gradient decreases (or remains below a certain threshold), this indicates convergence towards an optimum.

To understand the convergence properties of the FedFA framework, let's break down the gradient's behavior and its associated norms. Given the global gradient at iteration $t$ as:
\[ 
\lVert \nabla f_{G,\text{w/ scaled}}^t \rVert
\]
We can express it as an aggregation of the scaled gradients from each client:
\[ 
\lVert \nabla \left( \sum_{c\in{C}_{sel}} p_c \times \nabla f_c^{t,\mathrm{scaled}}) \right) \rVert
\]
The scaling factor $ \alpha_c^t $ alters the gradient norm:
\[
\lVert \nabla f_{G,\text{w/ scaled}}^t \rVert \leq \text{max}_{c\in{C}_{sel}}\{\alpha_c^t\} \times \lVert \nabla f_{G,\text{w/o scaled}}^t \rVert
\]
Using the bound from the unscaled case:
\[
\lVert \nabla f_{G,\text{w/ scaled}}^t \rVert \leq \text{max}_{c\in{C}_{sel}}\{\alpha_c^t\} \times \frac  {2L}{2+ \mu \eta}
\]
With the scaled full batch gradient descent update, the difference becomes:
\[
\lVert \omega_{t+1} - \omega_t \rVert = \eta \lVert \nabla f_{G,\text{w/ scaled}}^t \rVert
\]
Substituting the bound for the scaled gradient norm:
\[
\lVert \omega_{t+1} - \omega_t \rVert \leq \eta \times \text{max}_{c\in{C}_{sel}}\{\alpha_c^t\} \times \frac{2L}{2+ \mu \eta}
\]
Therefore, the convergence rate is:
\[
\mathcal{O} \left( \text{max}_{c\in{C}_{sel}}\left\{ \frac{\frac{1}{m} \sum_{\kappa \in {C}_{sel}} \left\| \omega_{95\%, \kappa} \right\|}{\left\| \omega_{95\%, c} \right\|} \right\} \times \frac{2L \eta}{2+ \mu \eta} \right) 
\]
This suggests that the convergence of FedFA is sensitive to the learning rate, scaling factors, and the inherent attributes of the loss function. It is crucial to recognize that the convergence rate usually fluctuates over time based on the selection of the learning rate \(\eta\), which could be either constant or adaptive, tending towards zero as the number of iterations \(t\) increases.
\newpage
\clearpage
\section{Scale Variations According to Heterogeneous Architectures} 
\label{apdx: scale variation}
\subsection{Introduction to Scale Variations in Federated Learning}
In FL, fair aggregation is challenging when local models' scales differ, a situation exacerbated by employing heterogeneous network architectures. This implies the necessity for scalable aggregation techniques. In this section, we empirically demonstrate scale variations across heterogeneous network architectures. 

\subsection{Empirical Analysis}
We employed three types of architectures: Pre-ResNet, MobileNetV2, and EfficientNetV2, varying in depth and width as depicted in Table~\ref{figure: model archi 1}. Detailed training conditions are in Table~\ref{figure: test condition}. To determine the depth $d_k$ and width $w_k$, we refer to the configuration in the second column of Table~\ref{figure: scalevariation_table}.

\subsubsection{Visualization of Weight Distributions}
\begin{figure}[ht]
    \centering
    \includegraphics[width=9 cm]{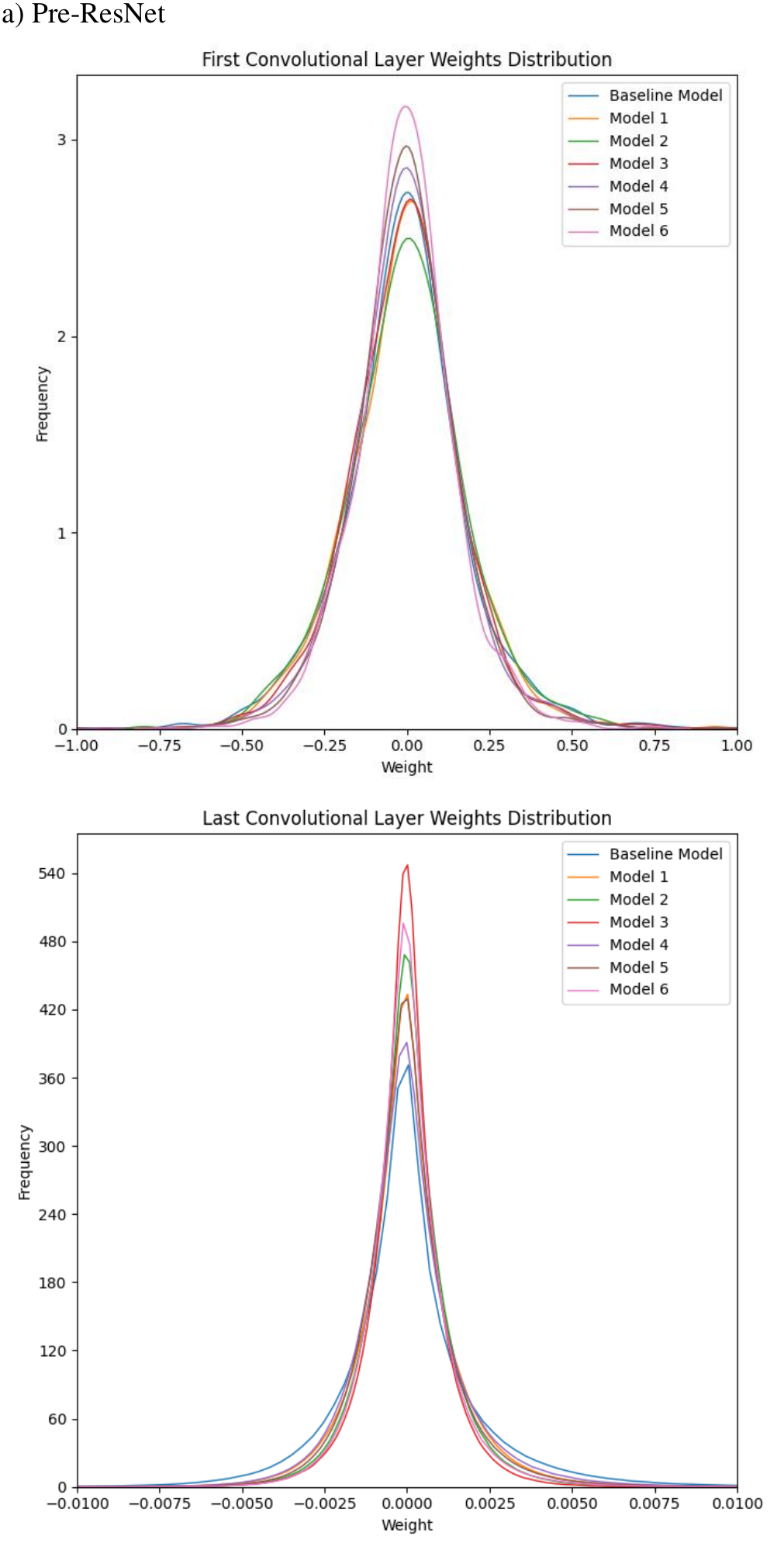}
    \caption{Weights distributions across different architectures of the first (left) and the last layers (right) for Pre-ResNet.}
    \label{fig: scale_variation_graph_1}
\end{figure}

\begin{figure}[ht]
    \centering
    \includegraphics[width=9 cm]{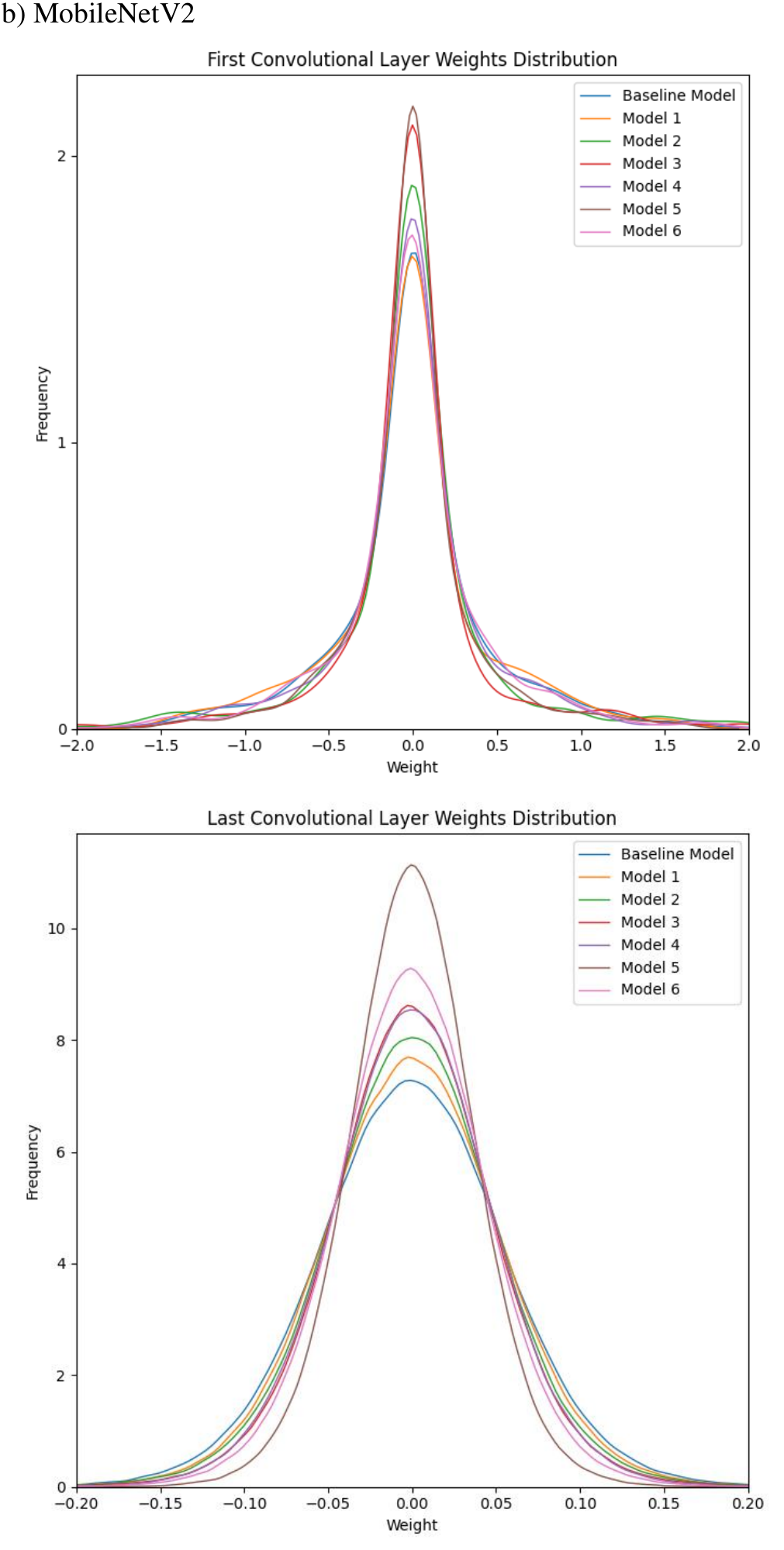}
    \caption{Weights distributions across different architectures of the first (left) and the last layers (right) for MobileNetV2.}
    \label{fig: scale_variation_graph_2}
\end{figure}

\begin{figure}[ht]
    \centering
    \includegraphics[width=9 cm]{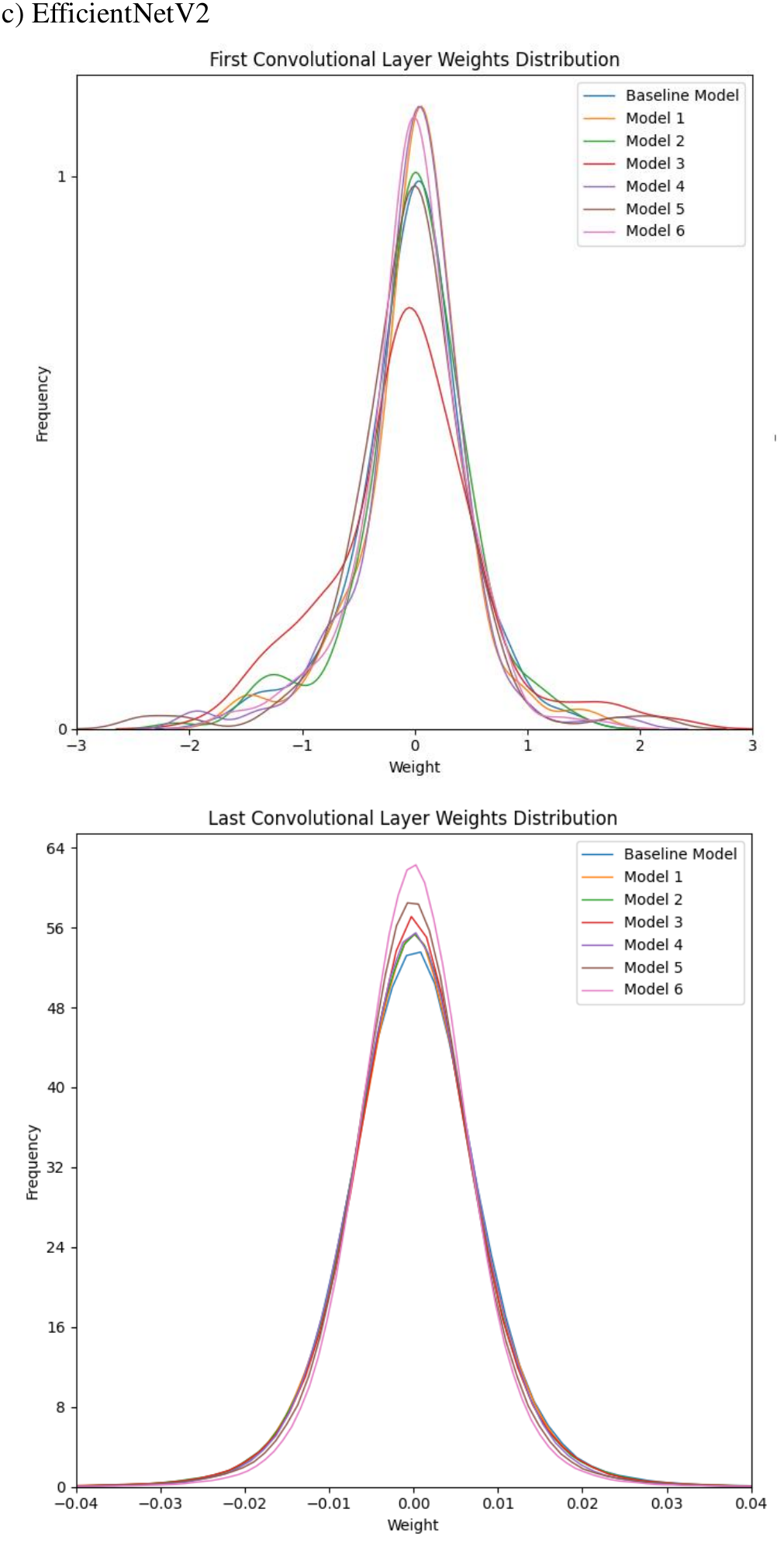}
    \caption{Weights distributions across different architectures of the first (left) and the last layers (right) for EfficientNetV2.}
    \label{fig: scale_variation_graph_3}
\end{figure}

For each model type, from the Baseline Model and Models 1 to 6, we flattened and vectorized all weights in the first and last convolutional layers to visualize the weight distributions as shown in Figures~\ref{fig: scale_variation_graph_1}, ~\ref{fig: scale_variation_graph_2}, and ~\ref{fig: scale_variation_graph_3}. These figures reveal unique weight distributions for each model, corresponding to their complexity.

\subsubsection{Quantifying Scale Variations}
We set the Baseline model $M$ to quantify scale variations with the minimum depth and width. In a single convolutional layer $l$ in the baseline model, there are $N_{out}^{M^{(l)}}$ filters, each composed of $N_{In}^{M^{(l)}}$ weight maps $M_i^{(l)}$, where $i$ is the index of each weight map.

\begin{table}[ht]
    \centering
    \caption{According to network complexities, each network has a distinct scale in its weights, leading to scale variations across heterogeneous network architectures.}
    \includegraphics[width=\linewidth]{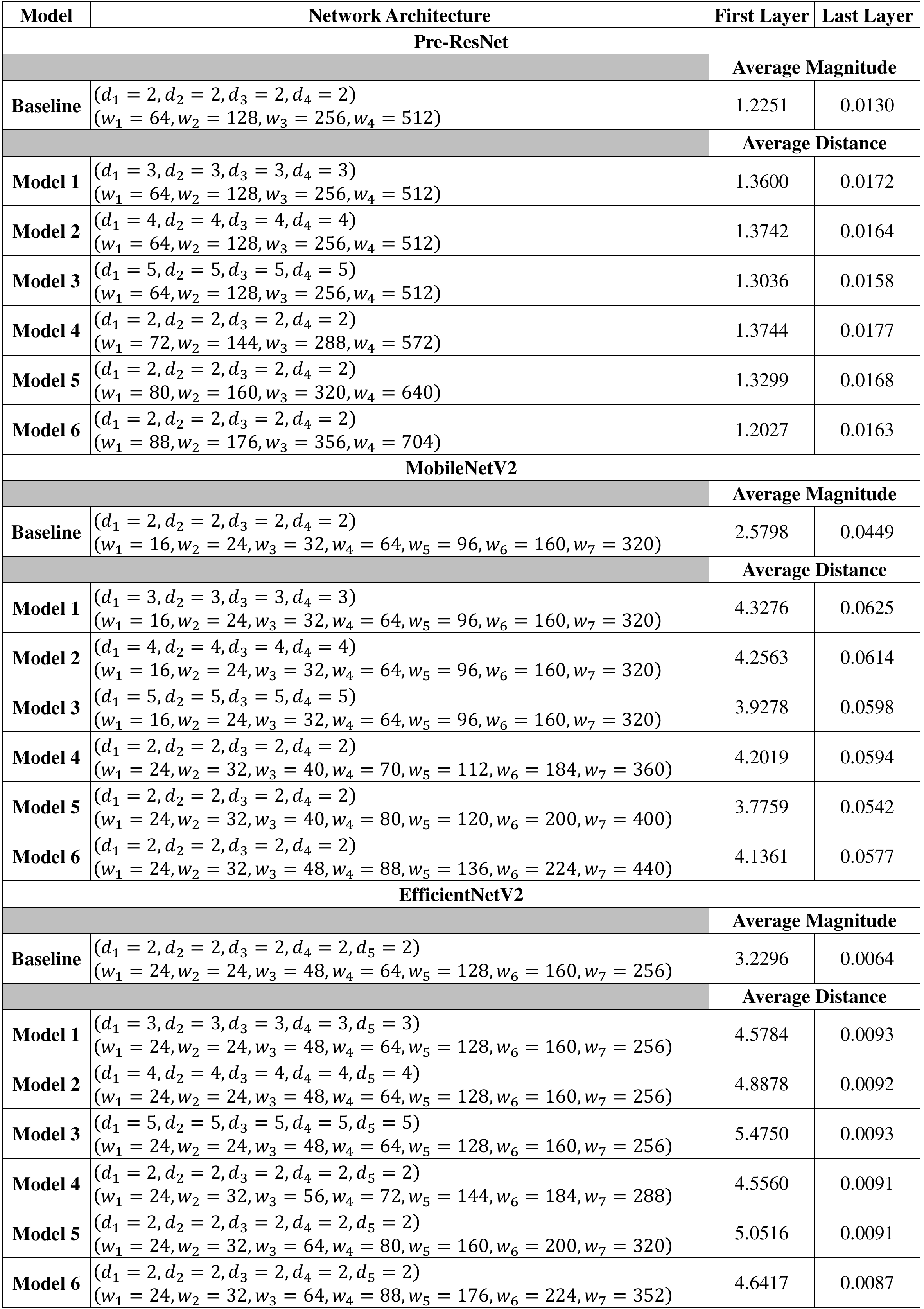}
    \label{figure: scalevariation_table}
\end{table}

We first examine the average magnitude:
\[
\text{Average Magnitude} = \frac{1}{N_{out}^{M^{(l)}} \times N_{In}^{M^{(l)}}}\sum_i \| M_i^{(l)} \|_1
\]
calculated as the averaged L1 Norms of weights in each weight map.

Next, we calculate the average distance from Models 1 to 6, separately from the Baseline model. To compare two networks of different complexities in width, we adopt the structured contiguous pruning concepts from HeteroFL~\cite{diao2020heterofl} and NeFL~\cite{kang2023nefl}, utilizing common parts of different layers. The pruned layer $l_k$ of Model $k$ is represented using indexing: $M_k^{(l_k)}[:N_{out}^{M^{(l)}},:N_{In}^{M^{(l)}}]$.

Average distance is calculated by:
\[
\text{Average Distance} = \frac{1}{N_{out}^{M^{(l)}} \times N_{In}^{M^{(l)}}}\sum_i \| M_i^{(l)} - M_k^{(l_k)}[:N_{out}^{M^{(l)}},:N_{In}^{M^{(l)}}] \|_1
\]

The average magnitudes of Baseline models and distances from the Baseline model to each Model $k$ are presented in Table~\ref{figure: scalevariation_table}. Comparing distances from Baseline models to each Model $k$, we observe variations according to network complexities, influenced by varying depth and width. Specifically, in Pre-ResNet, the distances are $0.98-1.36$ times greater than the Baseline models' average magnitude of weights. For MobileNetV2, these distances range from $1.20-1.68$ times the Baseline's average magnitude, and for EfficientNetV2, they are $1.35-1.70$ times greater. These findings empirically show the scale variations linked to network complexities and highlight the necessity for a scalable aggregation method in our FedFA framework.


\newpage
\clearpage
\section{Analysis of Scaling factors in HeteroFL}
\label{apdx: HeteroFL}
\subsection{Impact of Scaling Factors on Batch Normalization}
In the HeteroFL framework~\cite{diao2020heterofl}, scaling factors are utilized as part of the network to compensate for scale variations due to heterogeneous architectures during the training phase. These factors are decided based on network complexities. However, understanding the effect of scaling factors on gradients during the Batch Normalization (BN) process is crucial. We consider two simple model outputs: $\hat{y_1} = \sum_i a_i x_{i} + b$ without scaling, and $\hat{y_2} = \alpha (\sum_i a_i x_{i} + b)$ with scaling. If we apply the BN process to $\hat{y_1}$:
\[ BN(\hat{y_1}) = \frac{\hat{y_1} - \mu_1}{\sqrt{\sigma_1^2 + \epsilon_1}} \]
For $\hat{y_2}$, we adjust the mean ($\mu$) and variance ($\sigma^2$) by the scaling factor $\alpha$:
\[ \mu_2 = \alpha \mu_1, \quad \sigma_2^2 = \alpha^2 \sigma_1^2 \]
\[ BN(\hat{y_2}) = \frac{\hat{y_2} - \mu_2}{\sqrt{\sigma_2^2 + \epsilon_2}} 
= \frac{(\alpha\hat{y_1} - \alpha\mu_1)}{\sqrt{\alpha^2\sigma_1^2 + \epsilon_2}} 
\simeq \frac{\alpha (\hat{y_1} - \mu_1)}{\alpha \sqrt{\sigma_1^2 + \epsilon_2}} \]
The BN of the scaled output $\hat{y_2}$ simplifies to:
\[ \simeq \frac{\hat{y_1} - \mu_1}{\sqrt{\sigma_1^2 + \epsilon_2}} \]
\[ \Rightarrow BN(\hat{y_1}) \simeq BN(\hat{y_2}) \]
For the quadratic loss function $L = \frac{1}{2} (y - BN(\hat{y}))^2$, the gradients with respect to coefficients $a_i$ can be calculated for both outputs. \\
For $\hat{y_1}$ and its loss function $L_1$:
\[ \frac{\partial L_1}{\partial a_i} = \frac{\partial L_1}{\partial BN(\hat{y_1})} \times \frac{\partial BN(\hat{y_1})}{\partial \hat{y_1}} \times \frac{\partial \hat{y_1}}{\partial a_i} \]
For $\hat{y_2}$ and its loss function $L_2$, using the equivalence of the BN outputs and considering the scaling effect in the derivative:
\begin{equation*}
    \begin{aligned}
        \frac{\partial L_2}{\partial a_i} = \frac{\partial L_2}{\partial BN(\hat{y_2})} \times \frac{\partial BN(\hat{y_2})}{\partial \hat{y_2}} \times \frac{\partial \hat{y_2}}{\partial a_i} 
        &\simeq \frac{\partial L_1}{\partial BN(\hat{y_1})} 
        \times \frac{\partial BN(\hat{y_1})}{\alpha\partial \hat{y_1}}
        \times \frac{\alpha\partial \hat{y_1}}{\partial a_i} \\ 
        &=
        \frac{\partial L_1}{\partial BN(\hat{y_1})} \times \frac{\partial BN(\hat{y_1})}{\partial \hat{y_1}} \times \frac{\partial \hat{y_1}}{\partial a_i} = \frac{\partial L_1}{\partial a_i}
    \end{aligned}    
\end{equation*}

\subsection{Summary: Negation of Scaling Factor Effect}
This analysis demonstrates that the scaling factor $\alpha$ applied in $\hat{y_2}$ does not persist through the BN process. The equivalence in the BN outputs for both scaled and unscaled models leads to negating the $\alpha$ effect in the gradient computation. Consequently, applying scaling factors during the training may not adequately compensate for differences in model scales if networks have the BN layers, highlighting the need for another approach to managing scale variations.

\end{document}